\documentclass[10pt,twocolumn,letterpaper]{article}

\usepackage{iccv}              

\definecolor{iccvblue}{rgb}{0.21,0.49,0.74}
\usepackage[pagebackref,breaklinks,colorlinks,allcolors=iccvblue]{hyperref}

\usepackage{subfiles}
\usepackage{makecell}
\usepackage{arydshln}

\newcommand\blfootnote[1]{%
    \begingroup
    \renewcommand\thefootnote{}\footnote{#1}%
    \addtocounter{footnote}{-1}%
    \endgroup
}

\newcommand{\FIG}[1]{\cref{#1}}
\newcommand{\TABLE}[1]{\cref{#1}}
\newcommand{\EQUA}[1]{\cref{#1}}
\newcommand{\SECTION}[1]{\cref{#1}}

\newcommand{\MODEL}{Unnamed Model}

\newcommand{\xmark}{$\times$}
\newcommand{\xm}{\xmark}
\newcommand{\cm}{\checkmark}

\newcommand{\ETAL}{{\emph{et al.}}}

\newcommand{\IE}{{\emph{i.e.}}}

\newcommand{\PECNET}{PECNet}
\newcommand{\PECNETCITE}{\cite{mangalam2020not}}

\newcommand{\TRAJECTRONPPCITE}{\cite{salzmann2020trajectron}}
\newcommand{\YNET}{$\textsf{Y}$-net}
\newcommand{\YNETCITE}{\cite{mangalam2020s}}

\newcommand{\AGENTFORMERCITE}{\cite{yuan2021agentformer}}

\newcommand{\VMODEL}{V$^2$-Net}

\newcommand{\VCITE}{\cite{wong2022view}}

\newcommand{\MUSEVAE}{MUSE-VAE}
\newcommand{\MUSEVAECITE}{\cite{lee2022muse}}

\newcommand{\EVMODEL}{E-V$^2$-Net}

\newcommand{\EVCITE}{\cite{wong2023another}}
\newcommand{\EQMOTION}{EqMotion}
\newcommand{\EQMOTIONCITE}{\cite{xu2023eqmotion}}
\newcommand{\LED}{LED}
\newcommand{\LEDCITE}{\cite{mao2023leapfrog}}
\newcommand{\FLOWCHAIN}{FlowChain}
\newcommand{\FLOWCHAINCITE}{\cite{maeda2023fast}}
\newcommand{\IMP}{IMP}
\newcommand{\IMPCITE}{\cite{shi2023representing}}
\newcommand{\SCMODEL}{SocialCircle}
\newcommand{\SCCITE}{\cite{wong2023socialcircle}}

\newcommand{\LGTRAJ}{LG-Traj}
\newcommand{\LGTRAJCITE}{\cite{chib2024lg}}

\newcommand{\UPDD}{UPDD}
\newcommand{\UPDDCITE}{\cite{liu2024uncertainty}}
\newcommand{\SCPMODEL}{SocialCircle+}
\newcommand{\SCPCITE}{\cite{wong2024socialcircle+}}
\newcommand{\UEN}{UEN}
\newcommand{\UENCITE}{\cite{su2024unified}}
\newcommand{\MSTIP}{MS-TIP}
\newcommand{\MSTIPCITE}{\cite{chib2024ms}}
\newcommand{\SMEMO}{SMEMO}
\newcommand{\SMEMOCITE}{\cite{marchetti2024smemo}}
\newcommand{\DICE}{Dice}
\newcommand{\DICECITE}{\cite{choi2024dice}}
\newcommand{\PPT}{PPT}
\newcommand{\PPTCITE}{\cite{lin2024progressive}}
\newcommand{\AFFLN}{AgentFormer-FLN}
\newcommand{\AFFLNCITE}{\cite{xu2024adapting}}
\newcommand{\SOPERMODEL}{SoperModel}
\newcommand{\SOPERMODELCITE}{\cite{yang2025sopermodel}}

\renewcommand{\MODEL}{\emph{Resonance}}
\newcommand{\MODELSHORT}{\emph{Re}}
\newcommand{\SUPP}{Appendix}

\newcommand{\C}[1]{{\textcolor{iccvblue}{#1}}}
\newcommand{\X}[1]{{\textcolor[HTML]{FF4F4F}{#1}}}
\newcommand{\TODO}[1]{\colorbox{yellow}{TODO}}
\newcommand{\hatbf}[1]{\hat{\mathbf{#1}}}

\newcommand{\NEWHALFLINE}{\vspace{0.5em}\noindent}
\newcommand{\NOINDENT}{\noindent}

\def\paperID{7291} 
\def\confName{ICCV}
\def\confYear{2025}

\title{Resonance: Learning to Predict Social-Aware Pedestrian Trajectories\\ as Co-Vibrations}

\author{
    Conghao Wong\quad
    Ziqian Zou\quad
    Beihao Xia\quad
    Xinge You\\
Huazhong University of Science and Technology\\
{\tt\footnotesize
    conghaowong@icloud.com,
    ziqianzoulive@icloud.com,
    xbh\_hust@hust.edu.cn,
    youxg@mail.hust.edu.cn
}
}

\begin{document}
\maketitle

\begin{abstract}

    Learning to forecast trajectories of intelligent agents has caught much more attention recently.
    However, it remains a challenge to accurately account for agents' intentions and social behaviors when forecasting, and in particular, to simulate the unique randomness within each of those components in an explainable and decoupled way.
    Inspired by vibration systems and their resonance properties, we propose the \MODEL~(short for \MODELSHORT)~model to encode and forecast pedestrian trajectories in the form of ``co-vibrations''.
    It decomposes trajectory modifications and randomnesses into multiple vibration portions to simulate agents' reactions to each single cause, and forecasts trajectories as the superposition of these independent vibrations separately.
    Also, benefiting from such vibrations and their spectral properties, representations of social interactions can be learned by emulating the resonance phenomena, further enhancing its explainability.
    Experiments on multiple datasets have verified its usefulness both quantitatively and qualitatively.

\end{abstract}

\section{Introduction}

Trajectory prediction\blfootnote{
    Codes at \url{https://github.com/cocoon2wong/Re}.
} is a time series forecasting \cite{alahi2016social} task.
It aims at forecasting agents' behaviors in the form of trajectories by considering their statuses and potential interactive behaviors.
It can be used in applications like tracking \cite{fernando2018soft,pellegrini2009youll,saleh2020artist}, navigation \cite{alahi2017learning,chai2019multipath,chen2022scept}, and autonomous driving \cite{kim2017probabilistic,lee2017desire,phong2024truly}.
Pedestrian trajectories could be affected by multiple causes \cite{sadeghian2019sophie,liang2019peeking}.
Great efforts have been made to model agents' trajectories \cite{alahi2016social,mangalam2020not,mangalam2020s} and social behaviors \cite{xie2024pedestrian,marchetti2024smemo,lv2023skgacn} when forecasting, as well as their \emph{randomness} \cite{gupta2018social,liu2024uncertainty,lee2022muse} to make decisions under specific contexts.

However, most current approaches for representing randomnesses within trajectories could not distinguish and separately describe those triggered by different trajectory causes (we call \emph{determinants}).
Whether GAN-based \cite{rossi2021human,kosaraju2019social} or VAE-based \cite{yue2022human,lee2022muse,xu2022socialvae}, these methods model such randomnesses \emph{all-in-one}, then sampling features to generate randomized predictions.
In actual scenes, taking pedestrians as examples, they may present different random preferences targeting different trajectory determinants \cite{chen2021personalized,wong2021msn}.
Like someone introverted but rushing to meet a due date might more randomly choose their intentions but avoid social interactions.
In contrast, someone extroverted but with a fixed route preference might show more randomness in social choices than intentions.
Therefore, not distinguishing between different randomnesses leads to difficulty in capturing varying preferences of agents, also failing to ensure the causality \cite{granger1969investigating} of randomnesses in predictions, as well as their explainability.
Thus, our goal is to find a method that could decouple such randomnesses when forecasting.

\begin{figure}[t]
    \centering
    \includegraphics[width=1.0\linewidth]{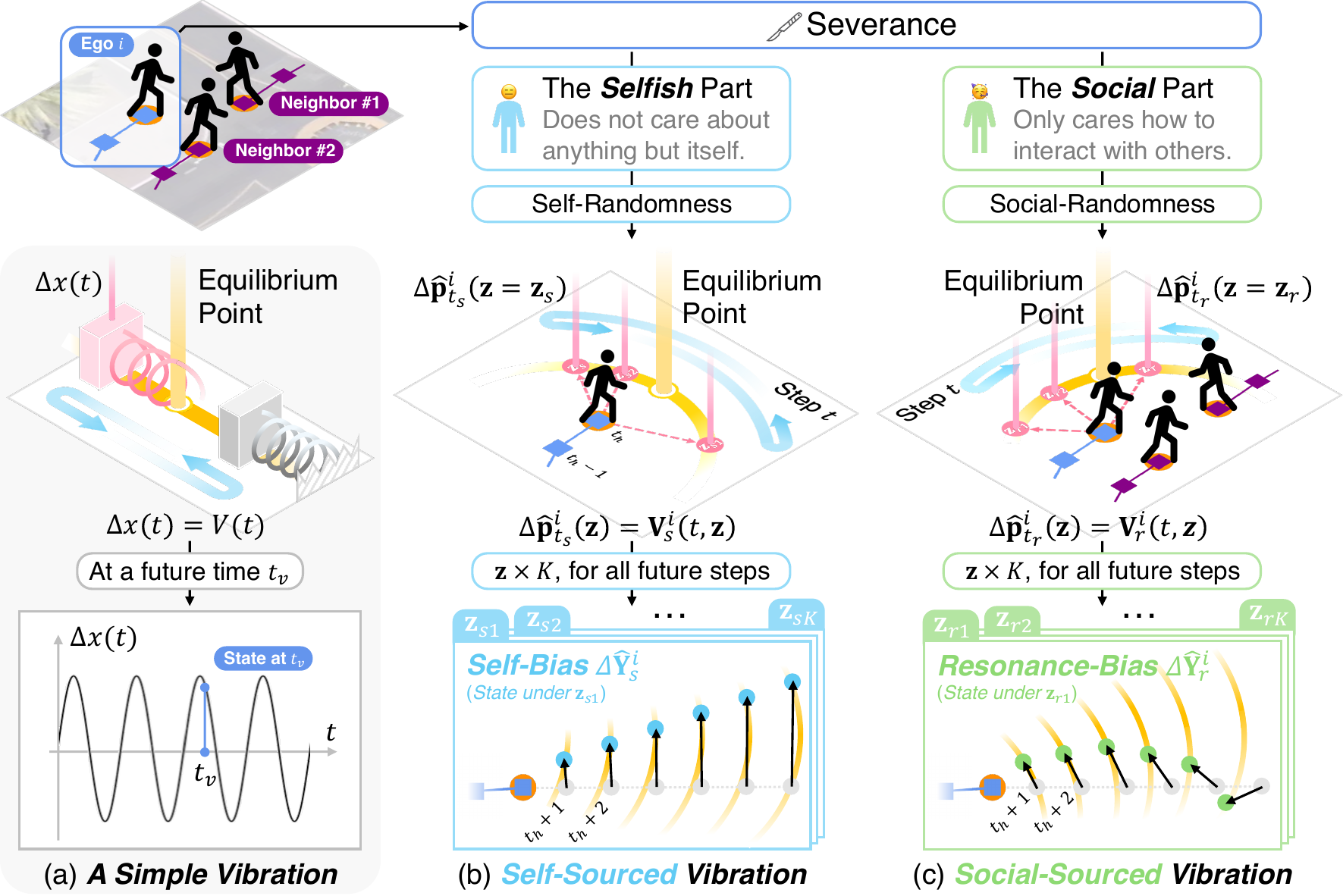}
    \caption{
        Motivation illustration (I).
        We predict trajectories as the superposition of multiple vibrations.
        At any prediction step $t$, each vibration describes the unique randomness of an ego agent under the only cause, whose state is determined by the noise variable $\mathbf{z}$.
    }
    \label{fig_intro}
\end{figure}

\begin{figure}[t]
    \centering
    \includegraphics[width=1.0\linewidth]{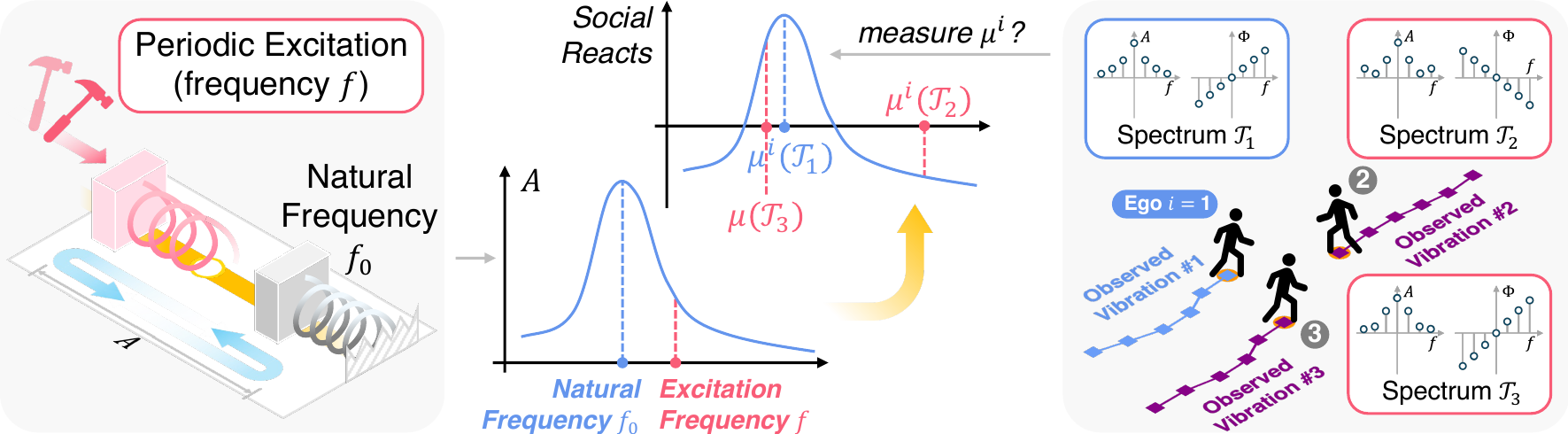}
    \caption{
        Motivation illustration (II).
        Analogous to the resonance of vibrations, we try to simulate social interactions in a resonance-like way by comparing spectral properties of egos' and all neighbors' (as excitations) vibrations, \IE, their observed trajectories.
    }
    \label{fig_intro_part2}
\end{figure}

\emph{EVERYTHING VIBRATES.}
As illustrated in \FIG{fig_intro} (a), vibrations are oscillations around equilibrium points.
They are always associated with specific frequencies and spectral properties.
Vibrations in linear systems are superpositionable, meaning that such a vibration could be decomposed into or superposed with other vibrations corresponding to each single excitation.
It is also fascinating that the resonance phenomena might appear as results of superposition, where a vibration system (with natural frequency $f_0$) may respond with a maximum amplitude when the frequency $f$ of an excitation vibration is the same as its natural frequency $f_0$, shown in the left of \FIG{fig_intro_part2}.
Once spectral properties of a vibration are determined, its state $V(t_v)$ can be inferred at any future time $t_v$ by solving differential equations.

Vibration theories characterize oscillatory motions.
Correspondingly, the randomnesses within agents' trajectories can somehow also be regarded as vibration systems, for which only a set of equilibrium points need to be assumed.
It is inline with our intuitions that randomnesses in trajectories always grow within specific ranges, rather than being completely random, applied to those in both intention changes or social behaviors.
For an ego $i$, a natural thought is to imitate vibrations, decomposing trajectory randomnesses at each given prediction step $t$ (irrelevant to $t_v$) as the superposition of $n$ vibrations $\{\mathbf{V}^i_n(t,t_v)\}_n$, corresponding to $n$ different trajectory determinants.
The states of these vibrations will be determined by random variables $\{\mathbf{z} = \mathbf{z}_n\}_n$ instead of some specific $t_v$, thus learning and forecasting such randomnesses and trajectories as the series of decoupled vibration states $\{\mathbf{V}^i_n(t, \mathbf{z}_n)\}_n$, better seen in \FIG{fig_intro}.

Benefiting from properties of vibrations, we can further represent egos' social behaviors and social randomnesses as the \emph{social resonance} of their own vibrations, just like resonance of vibrations, illustrated in \FIG{fig_intro_part2}.
For social interactions, as the saying describes, \emph{people are more inclined to hang out with those on a similar spectrum}.
We can assume that social interactions are associated with the spectral properties of different agents' vibrations, \IE, trajectories, and it could reach the maximum effects when their spectral properties satisfy some specific measurements.
This phenomenon could bring further explainable help if we use this resonance-like way to characterize social interactions, especially under our vibration-like trajectory prediction strategy.

The proposed \MODEL~(short for \MODELSHORT) model attempts to decompose randomnesses within trajectories into $n = 2$ parts, \emph{self-randomness} and \emph{social-randomness}, corresponding to \emph{self-sourced} and \emph{social-sourced} vibrations (\FIG{fig_intro} (b) and (c)), thus achieving the goal of distinguishing and separately forecasting each of those randomnesses, also infers social randomness as \emph{social resonance} between different vibrations, providing further explanations of social behaviors.

In summary, we contribute
(1) the ``vibration-like'' prediction strategy that simulates and decomposes the randomnesses in pedestrian trajectories as multiple vibrations according to different causes;
(2) the ``resonance-like'' representation of social interactions analogous to the resonance phenomena of vibrations;
and (3) experiments on multiple datasets have validated the effectiveness of the proposed \MODEL~model both qualitatively and quantitatively, especially the interpretability of predicted trajectories.

\section{Related Works}


\textbf{Trajectory Prediction and Social Interactions.}
Trajectory prediction is a time-series forecasting task.
Recently, researchers have employed recurrent neural networks \cite{jain2016structural,zhang2020social,quan2021holistic,rossi2021human,huang2021lstm,song2020pedestrian} to achieve such a goal.
Further, hierarchical predictions have also been made by adding destinations \cite{tran2021goal,mangalam2020not,rehder2018pedestrian,rehder2015goal,wong2021msn} or waypoint conditions \cite{chib2024enhancing,mangalam2020s,wong2022view}.
As for social interactions, former works \cite{mehran2009abnormal,pellegrini2009youll,luber2010people,anvari2015modelling} use Social Force \cite{helbing1995social} to describe agents' interaction context.
Then, Social Pooling~\cite{alahi2016social,gupta2018social}, attention mechanisms \cite{vemula2018social,mao2020history,shafiee2021introvert,li2021rain,ge2023causal} are introduced to measure the influence of different interaction participants.
Moreover, graph networks \cite{mohamed2020social,liu2021avgcn,huang2019stgat,duan2022complementary}, Transformers \cite{yu2020spatio,yuan2021agentformer,li2022graph,zhou2022hivt}, Diffusion models \cite{li2024bcdiff,mao2023leapfrog,bae2024singulartrajectory,liu2024uncertainty} and even large language models \cite{bae2024can,chib2024lg} are utilized to help represent social behaviors and forecast trajectories.
However, it is still difficult to tell from their predictions the effect of different trajectory-affecting causes, not to mention their separate randomnesses.

\NEWHALFLINE\textbf{Randomnesses in Trajectory Prediction.}
Randomnesses in future trajectories has also received attention when forecasting in recent years.
Gupta \ETAL~\cite{gupta2018social} first introduce Generative Adversarial Network (GAN) to generate multiple predicted trajectories that all meet social rules.
Then, more GAN-based prediction approaches \cite{sadeghian2019sophie,kosaraju2019social,kothari2023safety,rossi2021human,karras2019style} have been proposed to further enhance their randomizing capacities.
Due to the instability of training \cite{xu2022socialvae} of GANs, most newer approaches \cite{park2024t4p,neumeier2021variational,xu2022socialvae,lee2022muse} share Variational-Autoencoder-like (VAE-like) structures to simulate randomnesses in trajectories.
Although some researchers have attempted to encode differences in agents' preferences by constructing multiple latent spaces \cite{chen2021personalized,wong2021msn}, these approaches still could not distinguish such randomnesses of different trajectory determinants separately and uniquely.
This is the main concentration of the proposed \MODEL~model.

\section{Method}

\begin{table}
    \centering
    \footnotesize
    \begin{tabular}{p{1.7cm}l}
        \toprule
        Layers & Structures ($\sigma := \mathrm{ReLU}$, output shapes in ``[]'')\\

        \midrule

        $N_e$, $N_{e, l}$, $N_{r1}$ &
        $
            \left(\mbox{fc}(d/2, \sigma)\right)^2 \to
            \mbox{fc}(d/2, \tanh)
            \to [T_h, d/2]
        $. \\

        $N_{r2}$ &
        $
            \left(\mbox{fc}(d, \sigma)\right)^2 \to
            \mbox{fc}(d/2, \tanh)
            \to [d/2]
        $. \\
        
        $T_s$, $T_r$ &
        Transformer backbones (see \SECTION{sec_ID})
        $\to [T_h, d]$. \\

        $D_s$ &
        \makecell[l]{
            $
                \mbox{flatten}() \to 
                \mbox{fc}(dT_{way}, \sigma) \to
                \mbox{reshape}(T_{way}, d) 
            $ \\
            $
                \qquad\to \mbox{fc}(d, \sigma) \to
                \mbox{fc}(M)
                \to [T_{way}, M]
            $.
        } \\

        $D_r$ &
        \makecell[l]{
            $
                \mbox{flatten}() \to 
                \mbox{fc}(dT_f, \sigma) \to
                \mbox{reshape}(T_f, d) 
            $ \\
            $
                \qquad\to \mbox{fc}(d, \sigma) \to
                \mbox{fc}(M)
                \to [T_f, M]
            $.
        } \\

        \bottomrule
    \end{tabular}
    \caption{
        Detailed structures of layers used in \MODEL~model.
    }
    \label{tab_method}
\end{table}

\begin{figure}[t]
    \centering
    \includegraphics[width=1.0\linewidth]{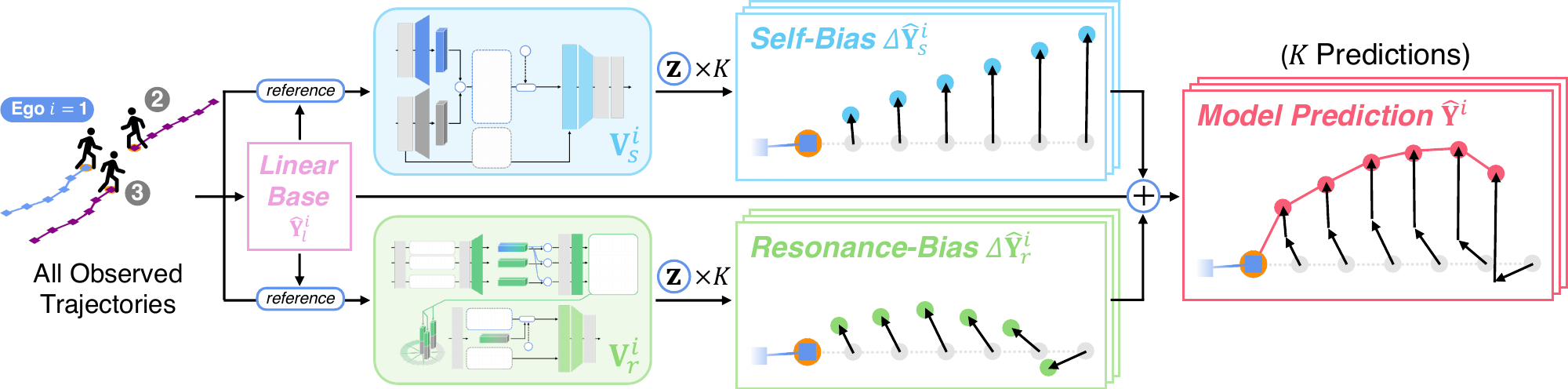}
    \caption{
        Method overview.
        Each prediction is obtained by superpositioning multiple trajectory biases (sampled from vibrations $\mathbf{V}^i_s(\mathbf{z})$ and $\mathbf{V}^i_r(\mathbf{z})$).
        They are supposed to own unique randomnesses and spectral properties upon linear base (reference points).
    }
    \label{fig_method_overview}
\end{figure}

\begin{figure}[t]
    \centering
    \includegraphics[width=1.0\linewidth]{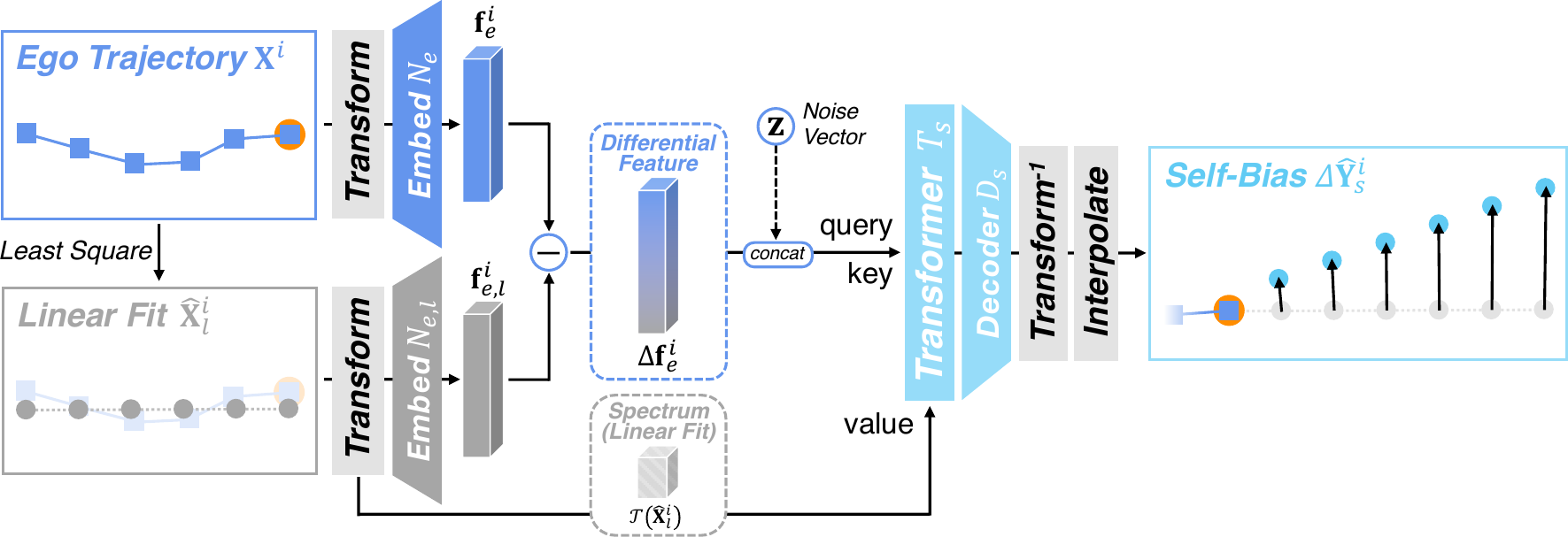}
    \caption{
        Computation pipeline of a self-bias $\Delta \hat{\mathbf{Y}}^i_s$ for agent $i$.
    }
    \label{fig_method_self}
\end{figure}

\textbf{Problem Settings.}
Denote 2D position (coordinate) of the $i$th ego agent at time step $t$ as $\mathbf{p}^i_t = (x^i_t, y^i_t)^\top$, given $t_h$ discrete observation steps and the consequential $t_f$ prediction steps (both with interval $\Delta t$), we aim at forecasting each agent $i$ ($1 \leq i \leq N_a$) one or more future trajectories $\hat{\mathbf{Y}}^i = (\hat{\mathbf{p}}^i_{t_h + 1}, ..., \hat{\mathbf{p}}^i_{t_h + t_f})^\top \in \mathbb{R}^{t_f \times 2}$ according to its observation $\mathbf{X}^i = ({\mathbf{p}}^i_{1}, ..., {\mathbf{p}}^i_{t_h})^\top \in \mathbb{R}^{t_h \times 2}$ and all neighbors' historical trajectories $\mathcal{X}^{-i} = \{\mathbf{X}^j | 1 \leq j \leq N_a, j \neq i\}$.

The core of the proposed \MODEL~(\MODELSHORT)~model is to simulate trajectories and their randomnesses as multiple vibrations, whose states are determined by the noise variable $\mathbf{z}$.
By sampling vectors $\{\mathbf{z}_s, \mathbf{z}_r\}$ (details in \EQUA{eq_selfbias,eq_rebias}), we have several \textbf{trajectory biases}, including linear base $\hat{\mathbf{Y}}^i_l$ as reference, self-bias $\hat{\mathbf{Y}}^i_s = \mathbf{V}^i_s (\mathbf{z} = \mathbf{z}_s)$ sampled from self-sourced vibration $\mathbf{V}^i_s (\mathbf{z})$, and resonance-bias $\Delta \hat{\mathbf{Y}}^i_r = \mathbf{V}^i_r (\mathbf{z} = \mathbf{z}_r)$ sampled from social-sourced vibration $\mathbf{V}^i_r (\mathbf{z})$.
Then, as shown in \FIG{fig_method_overview}, we have a prediction
\begin{equation}
    \label{eq_target}
    \hat{\mathbf{Y}}^i =
        \hat{\mathbf{Y}}^i_l + 
        \Delta \hat{\mathbf{Y}}^i_s +
        \Delta \hat{\mathbf{Y}}^i_r.
\end{equation}

\NOINDENT(i) \textbf{Linear Base.}
Intuitively, agents would prefer to maintain their motion status in the short term if there are no other distractions or intention changes.
It acts as the reference for all other vibrations.
We simulate these ``ideal'' equilibrium points via the linear least squares method.
Given the linear weight matrix $\mathbf{w}_l^i = (\mathbf{w}_l^{ix}, \mathbf{w}_l^{iy}) \in \mathbb{R}^{2 \times 2}$, we have the linear fit $\hat{\mathbf{X}}^i_l$ of agent $i$'s observed trajectory
\begin{align}
    \hat{\mathbf{X}}^i_l = \mathbf{A}_h \mathbf{w}_l^i,\quad
    \mbox{where}~\mathbf{A}_h = \left(\begin{matrix}
        1 & 1 & ... & 1 \\
        1 & 2 & ... & t_h \\
    \end{matrix}\right)^\top.
\end{align}
Linear weights $\mathbf{w}_l^i$ is resolved by minimizing $\Vert \hat{\mathbf{X}}^i_l - \mathbf{X}^i \Vert^2$:
\begin{equation}
    \mathbf{w}_l^i = \left(\mathbf{A}_h^\top \mathbf{A}_h\right)^{-1} \mathbf{A}_h^\top \mathbf{X}^i.
\end{equation}
Then, we can extrapolate the linear base $\hat{\mathbf{Y}}^i_l \in \mathbb{R}^{t_f \times 2}$ by
\begin{equation}
    \label{eq_linearbase}
    \hat{\mathbf{Y}}^i_l = \left(\begin{matrix}
        1 & 1 & ... & 1 \\
        t_h + 1 & t_h + 2 & ... & t_h + t_f \\
    \end{matrix}\right)^\top \mathbf{w}_l^i.
\end{equation}

\NEWHALFLINE
\noindent(ii) \textbf{Self-Sourced Vibration and Self-Bias.}
Self-sourced vibration is used to model future intentions and self-sourced randomness of ego agents, taking the linear base as the reference.
We use a differential structure (\FIG{fig_method_self}) to achieve this goal.
All computations are put in the frequency domain to capture its spectral properties, by first applying transform $\mathcal{T}$\footnote{
    Transform $\mathcal{T}$ may change shapes of inputs.
    For a 2D trajectory (define $m=2$) $\mathbf{X} \in \mathbb{R}^{t \times m}$, we represent the shape of its trajectory spectrum $\mathcal{T}(\mathbf{X})$ with the capitalization of the corresponding letter as ``$T \times M$''.
} on both the observed trajectory $\mathbf{X}^i$ and its linear fit $\hat{\mathbf{X}}^i_l$.
Two mirrored embedding networks $N_e$ and $N_{e, l}$ (see structures in \TABLE{tab_method}) will be applied to embed spectral representations $\mathbf{f}^i_e, \mathbf{f}^i_{e, l}$ accordingly.
Our concern lies in their \emph{differential feature} $\Delta \mathbf{f}^i_e$.
Denote feature dimensions as $d$, we have
\begin{gather}
    \label{eq_diff_encoding}
    \mathbf{f}^i_e = N_e \left(\mathcal{T}\left(\mathbf{X}^i\right)\right), \quad
    \mathbf{f}^i_{e, l} = N_{e, l} \left(\mathcal{T}\left(\hat{\mathbf{X}}^i_l\right)\right), \\
    \Delta \mathbf{f}^i_e = \frac{1}{2} \left(\mathbf{f}^i_e - \mathbf{f}^i_{e, l}\right)  \in \mathbb{R}^{T_h \times d/2}.
\end{gather}

Given noise variable $\mathbf{z} \sim \mathbf{N}(\mathbf{0}, \mathbf{I})$, we use a standard encoder-decoder structured Transformer \cite{vaswani2017attention}, named $T_{s}$, to encode the spectral feature $\mathbf{f}^i_s (\mathbf{z}) \in \mathbb{R}^{T_h \times d}$, which represents spectral properties of this vibration $\mathbf{V}^i_s (\mathbf{z})$.
The Transformer encoder computes self-attention over $T_h$ observed frequency portions within $\Delta \mathbf{f}^i_e$, learning the feature-level spectral differences upon reference points.
Its outputs serve as keys and queries in the Transformer decoder, while spectrum $\mathcal{T}(\hat{\mathbf{X}}^i_l) \in \mathbb{R}^{T_h \times M}$ serves as the target value to make sure that $\mathbf{f}^i_s (\mathbf{z})$ could represent future intention changes and randomness in a linear-trajectory-like way, also ensuring that such forecasted vibration occurs around linear base.
A state of vibration, \IE, self-bias, is represented by sampling $\mathbf{z} = \mathbf{z}_s$ ($\mathbf{z}_s \in \mathbb{R}^{T_h \times d/2}$, $\mathbf{z}_s = \mathbf{0}$ leads to equilibrium points):
\begin{equation}
    \label{eq_selfbias}
    \mathbf{f}^i_s (\mathbf{z} = \mathbf{z}_s) = T_s \left(
        \mathrm{Concat} \left(\Delta \mathbf{f}^i_e, \mathbf{z}_s\right),
        \mathcal{T}\left(\hat{\mathbf{X}}^i_l\right)
    \right).
\end{equation}

Then, a decoder MLP (named $D_s$, see \TABLE{tab_method}) is applied to decode features and predict self-biases.
Like previous waypoints-based approaches \cite{mangalam2020s}, self-bias is only assumed to \emph{roughly} reflect agents' trajectory changes on limited waypoints.
Thus, this decoder first flattens $\mathbf{f}^i_s (\mathbf{z}_s)$ on the last two dimensions ($T_h \times d$) and reshapes it to make sure forecasting features with a shape $T_{way} \times d$.
Then, a rough waypoint-bias trajectory $\Delta \hat{\mathbf{Y}}^i_{way} \in \mathbb{R}^{t_{way} \times 2}$ (with $t_{way}$ waypoints, whose temporal indices are equal-interval sampled) will be first forecasted after applying the inverse transform $\mathcal{T}^{-1}$.
Self-bias $\Delta \hat{\mathbf{Y}}^i_s \in \mathbb{R}^{t_f \times 2}$ is finally obtained by linearly interpolating (denoted as $I_l$) this waypoint-bias.
Formally,
\begin{equation}
    \label{eq_selfbiasdecoder}
    \Delta \hat{\mathbf{Y}}^i_s
        = I_l \left(
            \mathcal{T}^{-1} \left(
                D_s \left(
                    \mathbf{f}^i_s (\mathbf{z} = \mathbf{z}_s)
                \right)
            \right)
        \right).
\end{equation}

\begin{figure}[t]
    \centering
    \includegraphics[width=1.0\linewidth]{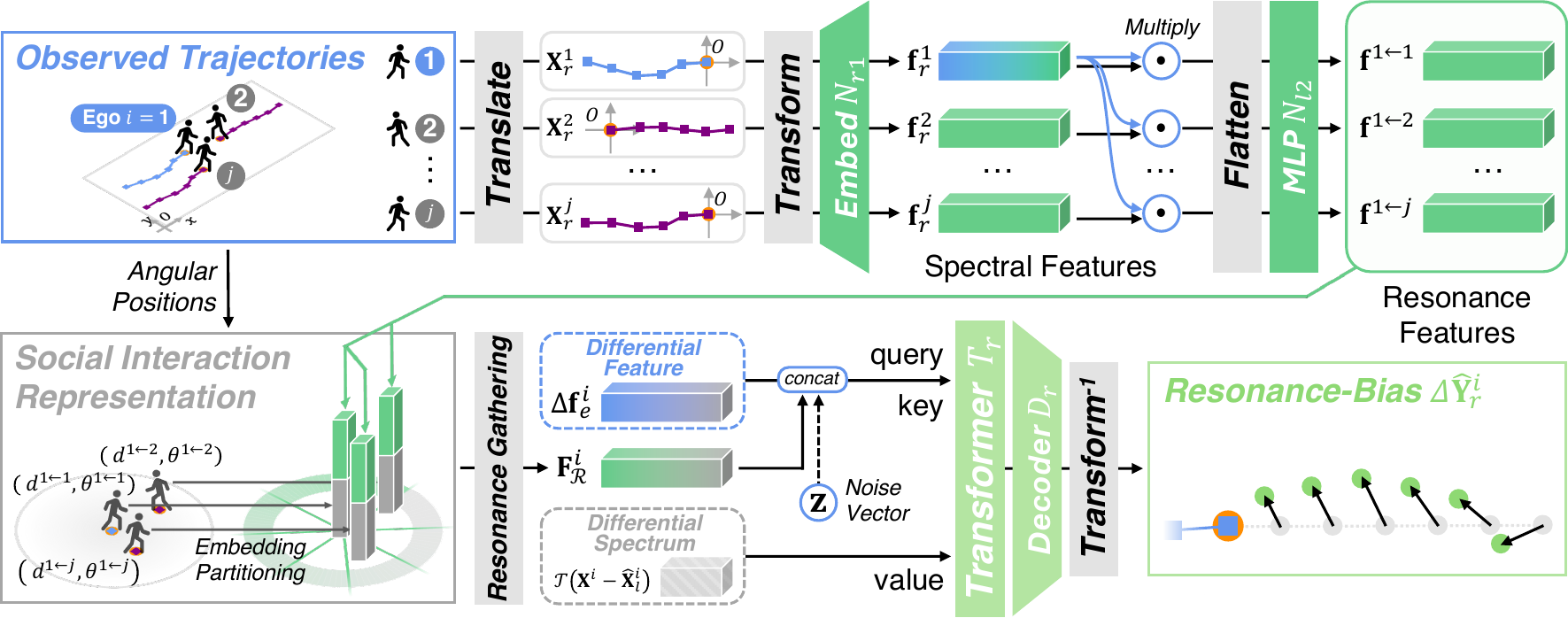}
    \caption{
        Computation pipeline of a resonance-bias $\Delta \hat{\mathbf{Y}}^i_r$ (ego $i$).
    }
    \label{fig_method_re}
\end{figure}

\noindent(iii) \textbf{Social-Sourced Vibration and Resonance-Bias (Re-Bias).}
As shown in \FIG{fig_method_re}, social-sourced vibration simulates future social behaviors and social randomness of egos when interacting with neighbors, also taking linear base as reference.
By considering trajectories as vibrations, social interactions could be simulated in a resonance-like manner by comparing spectral properties of different \emph{observed} vibrations, \IE, observed trajectories of different agents.
For ego $i$ and neighbor $j$, it can be described as how neighbor $j$ affects the vibration of ego $i$ (``\emph{natural frequencies}'' in \FIG{fig_intro_part2}) is supposed to be related to the spectral similarity with the vibration of neighbor $j$ (``\emph{excitation frequencies}'').

Before starting, we need to spectrally represent all these observed vibrations.
For all $u \in \{1, 2, ..., N_a\}$, we first translate every $\mathbf{X}^u$ so that they end at the origin.
This helps networks focus on the properties of trajectories themselves rather than the only positional relationships among agents.
Like \EQUA{eq_diff_encoding}, we use transform $\mathcal{T}$ and embedding network $N_{r1}$ (same structured as $N_e$) to represent each of them:
\begin{equation}
    \label{eq_relativeposition}
    \mathbf{f}^u_r = N_{r1} \left(\mathcal{T}\left(\mathbf{X}^u_r\right)\right),
    \quad\mbox{where}~~\mathbf{X}^u_r = \mathbf{X}^u - \mathbf{p}^u_{t_h}.
\end{equation}

For any agent pair $\{i, j\}$, \textbf{Resonance Feature} $\mathbf{f}^{i \leftarrow j} \in \mathbb{R}^{d/2}$ is then learned to compare spectral properties between vibrations (observed trajectories) of ego $i$ and neighbor $j$.
Given an encoder MLP $N_{r2}$ (structures in \TABLE{tab_method}), we have
\begin{equation}
    \mathbf{f}^{i \leftarrow j} = N_{r2} \left(
        \mathrm{Flatten}\left(
            \mathbf{f}^i_r \odot \mathbf{f}^j_r
        \right)
    \right).
\end{equation}
Here, $\odot$ represents the element-wise multiply.

$\mathbf{f}^{i \leftarrow j}$ only represents the potential resonance between agents $i$ and $j$.
Inspired by previous works \cite{wong2023socialcircle,wong2024socialcircle+}, we use an angle-based resonance gathering to gather resonance features of all neighbors.
It uses angle $\theta^{i \leftarrow j} = \mathrm{atan2}(
    \mathbf{p}^j_{t_h} - \mathbf{p}^i_{t_h}
)$ and distance $d^{i \leftarrow j} = \Vert
    \mathbf{p}^j_{t_h} - \mathbf{p}^i_{t_h}
\Vert$ to relatively locate each neighbor $j$, partitioning all neighbors into $N_\theta$ angular partitions, then gathering resonance features along with neighbors' positional information.
In the $n$th partition, we have the gathered resonance representation
\begin{equation}
    \label{eq_fij}
    \mathbf{F}^i_{\mathcal{R}} (n) = \mathbb{E}_{j \in \vartheta(n)} \left[
        \mathrm{Concat}\left(
            \mathbf{f}^{i \leftarrow j},
            \mathbf{f}^{i \leftarrow j}_p
        \right)
    \right] \in \mathbb{R}^d.
\end{equation}
Here, positional information $\mathbf{f}^{i \leftarrow j}_p$ is embedded with one linear layer ($d/2$ output units, $\tanh$) from $(d^{i \leftarrow j}, \theta^{i \leftarrow j})$, and $\vartheta(n) = \{
    j | 2\pi(n-1)/N_\theta \leq \theta^{i \leftarrow j} < 2\pi n/N_\theta
\}$ is the set of neighbors divided into the $n$th partition.
By stacking features in all partitions, we have the \textbf{Resonance Matrix}
\begin{equation}
    \mathbf{F}^i_{\mathcal{R}} = \left(
        \mathbf{F}^i_{\mathcal{R}} (1),
        \mathbf{F}^i_{\mathcal{R}} (2),
        ...,
        \mathbf{F}^i_{\mathcal{R}} (N_\theta)
    \right)^\top \in \mathbb{R}^{N_\theta \times d}
\end{equation}
to represent the overall spectral properties between the observed vibration of ego $i$ and those of all other neighbors, providing supervision for simulating and reproducing \emph{social resonance} in this forecasted social-sourced vibration.

Mirrored to $T_s$, re-biases are also sampled and forecasted through a Transformer $T_r$, which finally encodes a feature $\mathbf{f}^i_r (\mathbf{z}) \in \mathbb{R}^{T_h \times d}$.
Resonance matrix $\mathbf{F}^i_{\mathcal{R}}$ and differential feature $\Delta \mathbf{f}^i_e$ will be concatenated at the last dimension (zero-paddings required) and fed into the Transformer encoder, capturing possible \emph{social resonance} in the past.
Then, the differential spectrums between $\mathbf{X}^i$ and linear fit $\hat{\mathbf{X}}^i_l$ will be used as target values of queries in the Transformer decoder, learning how social behaviors and social randomness modify trajectories upon the linear fit during observation, inferring spectral properties of social-sourced vibration $\mathbf{V}^i_r (\mathbf{z})$ in the future.
Like \EQUA{eq_selfbias}, after sampling $\mathbf{z} = \mathbf{z}_r \in \mathbb{R}^{T_h \times d/2}$ (where $\mathbf{z}_r = \mathbf{0}$ also leads to equilibrium points), we have
\begin{equation}
    \label{eq_rebias}
    \mathbf{f}^i_r (\mathbf{z} = \mathbf{z}_r) = T_r \left(
        \mathrm{Concat}\left(
            \Delta \mathbf{f}^i_e,
            \mathbf{F}^i_{\mathcal{R}},
            \mathbf{z}_r
        \right),
        \mathcal{T}\left(
            \mathbf{X}^i - \hat{\mathbf{X}}^i_l
        \right)
    \right).
\end{equation}

By passing such feature through a decoder MLP ($D_r$ in \TABLE{tab_method}), which flattens $\mathbf{f}^i_r$ and resizes it to infer the spectral properties of such vibrations at all $t_f$ future steps, we have
\begin{equation}
    \Delta \hat{\mathbf{Y}}^i_r = \mathcal{T}^{-1} \left(
        D_r \left(
            \mathbf{f}^i_r (\mathbf{z} = \mathbf{z}_r)
        \right)
    \right).
\end{equation}

\noindent\textbf{Model Predictions and Training.}
\MODELSHORT~is trained end-to-end with the $\ell_2$ loss.
Under the \emph{best-of-$K$} \cite{gupta2018social} validation (using subscript $_k$, by sampling $\{\mathbf{z}_s, \mathbf{z}_r\}$ for $K$ times), we have
\begin{gather}
    \ell_2 \left(
        \mathbf{Y}^i,
        \left\{\hat{\mathbf{Y}}^i_k\right\}_{k=1}^{K}
    \right) = \min_{k} \left\Vert
        \mathbf{Y}^i - \hat{\mathbf{Y}}^i_k
    \right\Vert.
\end{gather}

\section{Experiments}

\begin{table*}[tbph]
    \centering
    \footnotesize
    \begin{tabular}{lcccccc}
        \toprule
        Method (ETH-UCY) & eth $\downarrow$ & hotel $\downarrow$ & univ $\downarrow$ & zara1 $\downarrow$ & zara2 $\downarrow$ & Average $\downarrow$ \\

        \midrule
        \MSTIP\MSTIPCITE~(2024) & 0.39/0.57 & 0.13/0.22 & 0.24/\C{0.40} & 0.20/0.34 & 0.17/0.29 & 0.22/0.36 \\
        \SMEMO\SMEMOCITE~(2024) & 0.39/0.59 & 0.14/0.20 & 0.23/0.41 & 0.19/0.32 & 0.15/0.26 & 0.22/0.35 \\
        \EQMOTION\EQMOTIONCITE~(2023) & 0.40/0.61 & 0.12/0.18 & 0.23/0.43 & 0.18/0.32 & \C{0.13}/0.23 & 0.21/0.35 \\
        \LED\LEDCITE~(2023) & 0.39/0.58 & \C{0.11}/0.17 & 0.26/0.43 & 0.18/0.26 & \C{0.13/0.22} & 0.21/0.33 \\
        Trajectron++\TRAJECTRONPPCITE~(2020) & 0.43/0.86 & 0.12/0.19 & \C{0.22}/0.43 & 0.17/0.32 & \C{0.12}/0.25 & 0.20/0.39 \\
        \LGTRAJ\LGTRAJCITE~(2024) & 0.38/0.56 & \C{0.11}/0.17 & 0.23/0.42 & 0.18/0.33 & 0.14/0.25 & 0.20/0.34 \\
        \PPT\PPTCITE~(2024) & 0.36/0.51 & \C{0.11/0.15} & \C{0.22/0.40} & 0.17/0.30 & \C{0.12/0.21} & 0.20/0.31 \\
        \EVMODEL\EVCITE~(2023) & 0.25/0.38 & \C{0.11}/0.16 & 0.23/0.42 & 0.19/0.30 & \C{0.13}/0.24 & \C{0.18}/0.30 \\
        AgentFormer\AGENTFORMERCITE~(2021) & 0.26/0.39 & \C{0.11/0.14} & 0.26/0.46 & \C{0.15/0.23} & 0.14/0.23 & \C{0.18}/0.29 \\
        \SCMODEL\SCCITE~(2024) & 0.25/0.38 & 0.12/\C{0.14} & 0.23/0.42 & 0.18/0.29 & \C{0.13/0.22} & \C{0.18}/0.29 \\
        \SCPMODEL\SCPCITE~(2024) & 0.25/0.39 & \C{0.10/0.15} & 0.24/0.42 & 0.18/0.28 & \C{0.13/0.22} & \C{0.18}/0.29 \\
        \YNET\YNETCITE~(2021) & 0.28/\C{0.33} & \C{0.10/0.14} & 0.24/0.41 & 0.17/\C{0.27} & \C{0.13/0.22} & \C{0.18/0.27} \\
        \UPDD\UPDDCITE~(2024) & \C{0.22}/0.42 & 0.17/0.30 & \C{0.14/0.28} & \C{0.16}/0.30 & 0.14/0.31 & \C{0.17}/0.32 \\
        
        \midrule
        \MODELSHORT~(Ours) & \C{0.23/0.35} & \C{0.10/0.15} & 0.24/0.41 & 0.17/0.29 & \C{0.13/0.22} & \C{0.17/0.28} \\ 

        \bottomrule
    \end{tabular}
    \begin{tabular}{lc}
        \toprule
        Method (SDD) & ADE/FDE $\downarrow$ \\

        \midrule
        \FLOWCHAIN\FLOWCHAINCITE~(2023) & 9.93/17.17 \\
        \IMP\IMPCITE~(2023) & 8.98/15.54 \\
        \LED\LEDCITE~(2023) & 8.48/11.36 \\
        \SMEMO\SMEMOCITE~(2024) & 8.11/13.06 \\
        \LGTRAJ\LGTRAJCITE~(2024) & 7.80/12.79 \\
        \YNET\YNETCITE~(2021) & 7.85/11.85 \\
        \UEN\UENCITE~(2024) & 7.30/10.40 \\
        \PPT\PPTCITE~(2024) & 7.03/10.65 \\
        \UPDD\UPDDCITE~(2024) & 6.59/13.90 \\
        \EVMODEL\EVCITE~(2023) & 6.57/10.49 \\
        \SCMODEL\SCCITE~(2024) & 6.54/10.36 \\
        \SCPMODEL\SCPCITE~(2024) & 6.44/\C{10.22} \\
        \MUSEVAE\MUSEVAECITE~(2022) & \C{6.36}/11.10 \\
        
        \midrule
        \MODELSHORT~(Ours) & \C{6.27/10.02} \\

        \bottomrule
    \end{tabular}
    \caption{
        Comparisons to other state-of-the-art methods on ETH-UCY (left) and SDD (right).
        Metrics are ``ADE/FDE'' (\emph{best-of-20}) in meters on ETH-UCY and in pixels on SDD.
        Lower metrics indicate better performance.
        \C{Blue} numbers mark the top 2 results on each set.
    }
    \label{tab_sota_ethucysdd}
  \end{table*}

\label{sec_expsettings}

\noindent\textbf{Datasets.}
(a) \textbf{ETH-UCY}~\cite{pellegrini2009youll,lerner2007crowds} comprises videos captured in pedestrian walking scenarios.
We use the \emph{leave-one-out} split under $\left(t_h, t_f, \Delta t\right) = (8, 12, 0.4)$ \cite{alahi2016social,zhang2019sr}.
(b) \textbf{Stanford Drone Dataset} (SDD)~\cite{robicquet2016learning} includes videos captured over the campus, covering different categories of agents.
We split 60\%/20\%/20\% to train/test/val, under $\left(t_h, t_f, \Delta t\right) = (8, 12, 0.4)$ \cite{liang2020simaug}.
(c) \textbf{NBA}~\cite{linou2016nba} includes trajectories (players and basketballs) captured during NBA games.
We set $\left(t_h, t_f, \Delta t\right) = (5, 10, 0.4)$, and randomly select 32,500/12,500/5,000 samples to train/test/val \cite{xu2022groupnet,xu2022remember}.

\NEWHALFLINE
\textbf{Metrics \& Implementation Details.}
\label{sec_ID}
We use the best Average/Final Displacement Error over $K$ trajectories (minADE$_K$/minFDE$_K$) to measure performance \cite{alahi2016social,gupta2018social}.
We abbreviate them as ``ADE'' and ``FDE''.
See definitions in \SUPP.
\MODELSHORT~is trained on one NVIDIA RTX 3090 with an Adam optimizer under a learning rate of 3e-4 and a batch size of 1000 for 200 epochs.
Transformers $T_s$ and $T_r$ have 8 attention heads.
$T_s$ has 4 encoder-decoder layers, while $T_r$ has 2.
Feature dimension ($d$) is set to 128.
Transform $\mathcal{T}$ is set to the discrete Haar transform \EVCITE.
Following previous works \cite{wong2022view,wong2023socialcircle}, we set $t_{way} = 4$ when $t_f = 12$, and $t_{way} = 3$ when $t_f = 10$, and set $N_\theta = t_h$.

\subsection{Quantitative Analyses}

\noindent\textbf{Comparisons to State-of-the-Arts.}
In \TABLE{tab_sota_ethucysdd}, it can be seen that \MODELSHORT~performs well on pedestrian datasets.
Compared to the newly published \SCPMODEL, \MODELSHORT~obtains about 8.0\% and 10.2\% better ADE and FDE on eth, and about 5.6\% and 3.7\% better average performance.
In addition, \MODELSHORT~improves about 15.6\% average FDE compared to \UPDD, which currently owns state-of-the-art performance on this set.
It improves about 2.6\% and 1.9\% ADE and FDE on the SDD compared to the \SCPMODEL, and about 19.6\% and 20.2\% performance compared to the newly published \LGTRAJ~driven by large language models.
\MODELSHORT~also obtained competitive results in the sports set NBA.
In \TABLE{tab_sota_nuscenes}, \MODELSHORT~outperforms GroupNet+CVAE by about 3.2\% and 0.9\% ADE accordingly.
Also, compared to \SCPMODEL~with the best FDE, \MODELSHORT~achieves a significant improvement of 9.3\% in FDE when $t_f = 5$, also 0.7\% better when $t_f = 10$.
These results indicate the superiority of \MODELSHORT~in forecasting pedestrian trajectories across diverse scenarios.\footnote{
    \MODELSHORT~mainly focuses on pedestrian trajectories.
    See results and discussions on the vehicle dataset nuScenes \cite{caesar2020nuscenes,caesar2019nuscenes} in \SUPP.
}

\begin{table}[tbp]
    \centering
    \footnotesize

        

    \begin{tabular}{lcc}
        \toprule
        Method (NBA) & $t_f = 5$ $\downarrow$ & $t_f = 10$ $\downarrow$ \\

        \midrule
        \PECNET\PECNETCITE~(2020) & 0.96/1.69 & 1.83/3.41 \\
        MemoNet\cite{xu2022remember}~(2022) & 0.71/1.14 & 1.25/1.47 \\
        GroupNet+NMMP\cite{xu2022groupnet}~(2022) & 0.69/1.08 & 1.25/1.80 \\
        GroupNet+CVAE\cite{xu2022groupnet}~(2022) & \C{0.62}/0.95 & \C{1.13}/1.69 \\
        \VMODEL\VCITE~(2022) & 0.69/0.96 & 1.28/1.68 \\
        \EVMODEL\EVCITE~(2023) & 0.68/0.93 & 1.26/1.64 \\
        \SCMODEL\SCCITE~(2024) & 0.67/0.90 & 1.18/1.46 \\
        \SCPMODEL\SCPCITE~(2024) & 0.65/\C{0.86} & 1.14/\C{1.37} \\
        
        \midrule
        \MODELSHORT~(Ours) & \C{0.60/0.78} & \C{1.12/1.38} \\

        \bottomrule
    \end{tabular}
    \caption{
        Comparisons to other state-of-the-art methods on NBA.
        Metrics reported are ``ADE/FDE'' in meters under \emph{best-of-20}.
    }
    \label{tab_sota_nuscenes}
  \end{table}


\begin{table*}[tbph]
    \centering
    \footnotesize
    \begin{tabular}{ccccccccccccccccccccccc}
        \toprule
        ID & ~$l$~~$\Delta_s$~$\Delta_r$ &
        eth $\downarrow$ &
        hotel $\downarrow$ &
        univ $\downarrow$ &
        zara1 $\downarrow$ &
        zara2 $\downarrow$ &
        ETH-UCY $\downarrow$ &
        SDD $\downarrow$ &
        NBA $\downarrow$ \\
        
        \midrule
        a1 & \cm~~\xm~~\xm & 0.53/1.10 & 0.22/0.43 & 0.59/1.24 & 0.47/1.01 & 0.36/0.77              & 0.434/0.910       & 15.72/32.30       &  4.55/8.72 \\
        a2 & \cm~~\cm~~\xm & 0.28/0.39 & 0.12/0.18 & 0.28/0.50 & 0.21/0.32 & 0.16/0.24              & 0.210/0.326       & 6.93/\C{10.26}    &  1.35/1.82 \\
        a3 & \xm~~\xm~~SC & 0.24/0.39 & 0.11/\C{0.16} & 0.29/0.53 & 0.18/0.31 & \C{0.13/0.23}       & 0.190/0.324       & 6.51/10.52        &  1.15/1.44 \\
        a4 & \xm~~\xm~~Re & 0.24/0.37 & \C{0.10/0.15} & 0.29/0.52 & 0.18/\C{0.29} & \C{0.13/0.22}   & 0.188/0.310       & 6.45/10.46        &  \C{1.12/1.38} \\
        a5 & \xm~~\cm~~SC & 0.24/0.38 & 0.11/\C{0.16} & 0.27/0.50 & 0.18/0.31 & 0.14/\C{0.23}       & 0.188/0.316       & 6.51/10.46        &  1.18/1.52 \\
        a6 & \xm~~\cm~~Re & 0.24/0.38 & \C{0.10/0.16} & 0.26/0.47 & 0.18/\C{0.30} & \C{0.13/0.22}   & \C{0.182/0.306}   & 6.40/\C{10.41}    &  \C{1.14/1.42} \\
        a7 & \cm~~\xm~~SC & \C{0.23/0.36} & 0.12/0.19 & 0.26/0.47 & 0.18/0.31 & 0.14/0.24           & 0.186/0.314       & 6.36/10.53        &  1.21/1.67 \\
        a8 & \cm~~\xm~~Re & 0.28/\C{0.36} & 0.11/0.18 & \C{0.25}/0.45 & 0.18/0.31 & 0.14/0.25       & 0.192/0.310       & \C{6.33}/10.45    &  1.21/1.64 \\
        a9 & \cm~~\cm~~SC & 0.24/0.37 & 0.11/0.18 & \C{0.25/0.44} & 0.18/\C{0.30} & 0.14/0.24       & 0.184/\C{0.306}   & 6.39/10.40        &  1.19/1.58 \\

        \midrule
        a0 & \cm~~\cm~~Re &
        \C{0.23/0.35} & \X{0.11/0.17} & \C{0.24/0.41} & \C{0.17/0.29} & \C{0.13/0.22} &
        \C{0.176/0.288 } &
        \C{6.27/10.02} &
        \X{1.17/1.55} \\

        \bottomrule
    \end{tabular}
    \caption{
        Ablation studies.
        ``$l$'', ``$\Delta_s$'', ``$\Delta_r$'' represent whether $\hat{\mathbf{Y}}^i_l$, $\Delta \hat{\mathbf{Y}}^i_s$, and $\Delta \hat{\mathbf{Y}}^i_r$ are superposed in \EQUA{eq_target} when training.
        ``Re'' and ``SC'' denote resonance gathering and \SCMODEL~\SCPCITE~when learning $\mathbf{V}^i_r$.
        Results marked in \X{Red} denote worse ones than the best variation.
    }
    \label{tab_ab}
  \end{table*}

\NEWHALFLINE\textbf{Ablation Studies.}
\label{sec_ablation}
\TABLE{tab_ab} reports several ablation results.\footnote{
    See ablation studies about $\{\mathcal{T}, t_{way}, N_\theta\}$ in \SUPP.
}
We can see that all vibration portions could effectively help on most pedestrian datasets.
However, \MODELSHORT~perform even worse when adding linear bases in the final predictions on the NBA dataset.
Variations a7 to a9 exhibit performance drops of up to 7.4\%/17.7\%, which are rarely seen in ETH-UCY and SDD.
Nevertheless, vibrations (self-biases and re-biases) could still help to improve the performance (variation a6), especially compared to the \SCMODEL~variation a5, which enhances 3.5\%/7.0\% ADE/FDE under the same conditions.
Though the assumption that linear base serve as the final reference points may not apply to NBA scenes, vibrations and their sampled trajectory-biases still work continuously, which proves the usefulness of both vibration-like prediction and resonance-like modeling strategies.

\subsection{Discussions}

\begin{figure}[b]
    \centering
    \includegraphics[width=1.0\linewidth]{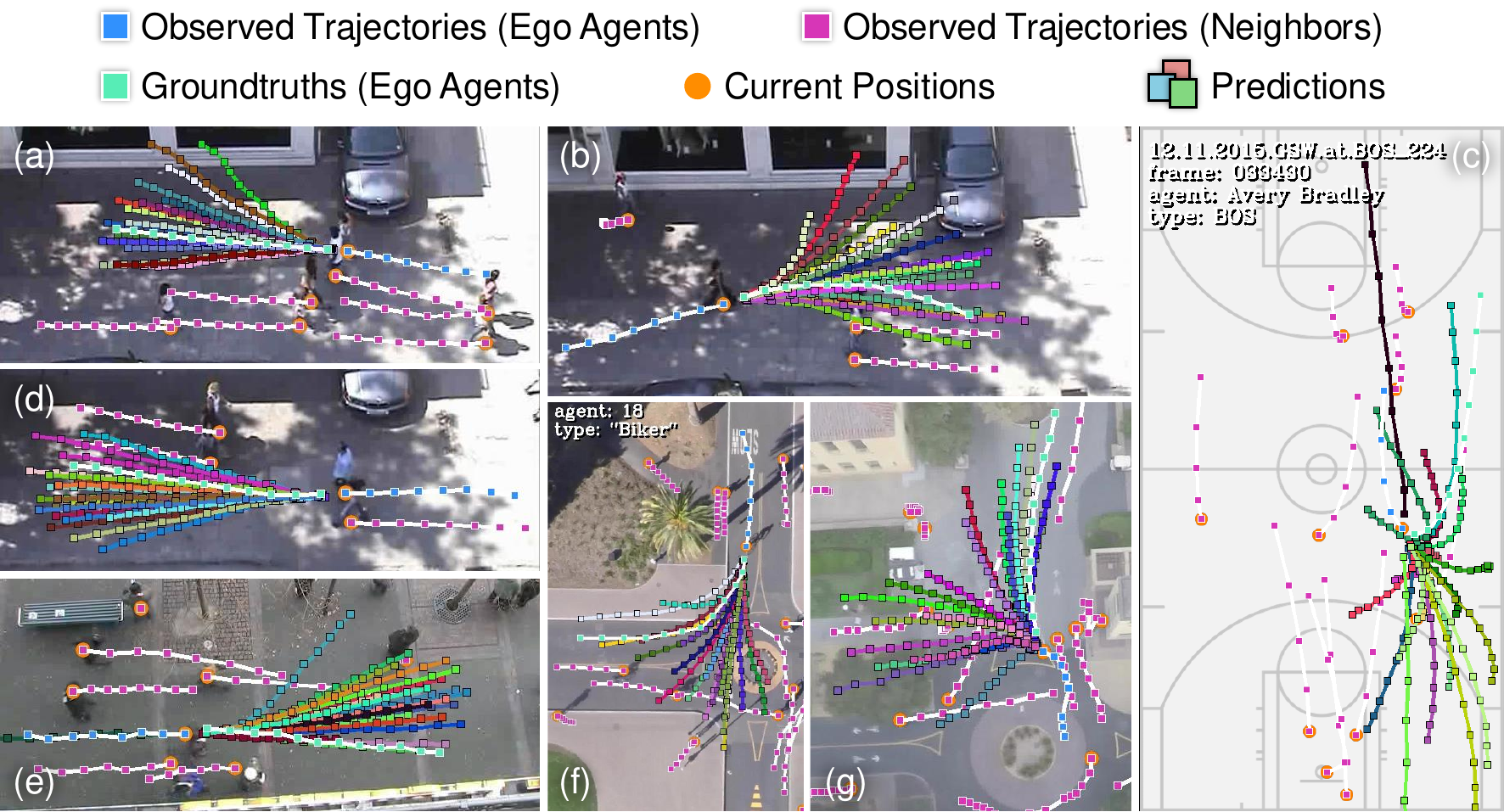}
    \caption{
        Visualized \MODELSHORT~predictions in different scenes (datasets).
    }
    \label{fig_vis}
\end{figure}

\noindent\textbf{Visualized Model Predictions.}
\FIG{fig_vis} visualizes trajectories predicted by \MODELSHORT.
It can be seen that interactive properties have been presented, like the ego in \FIG{fig_vis} (a) to avoid the down-coming neighbor or reserve paths for its companion.
It could also forecast bikers to cross the roundabout with different choices in \FIG{fig_vis} (f) and (g), and forecast NBA players under different offensive strategies in \FIG{fig_vis} (c).

\begin{figure*}[tbph]
    \centering
    \includegraphics[width=1.0\linewidth]{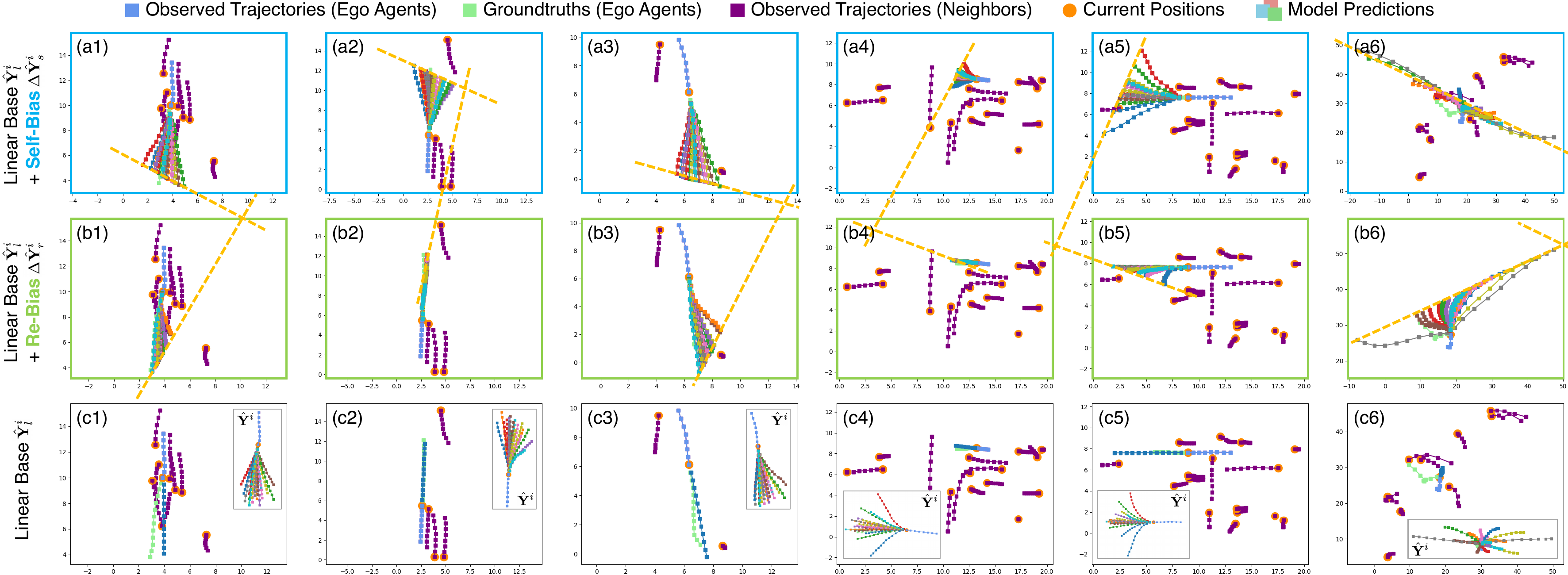}
    \caption{
        Visualized trajectory biases ($K=20$ samplings from each vibration).
        Subfigures ($x$1) to ($x$3) come from ETH-UCY scenes, ($x$4) to ($x$5) from SDD, and ($x$6) from NBA, where $x \in \{\mbox{a, b, c}\}$.
        Full predictions ($\hat{\mathbf{Y}}^i$) are drawn in additional gray boxes in the last row.
    }
    \label{fig_biases}
\end{figure*}

\begin{figure}[t]
    \centering
    \includegraphics[width=1.0\linewidth]{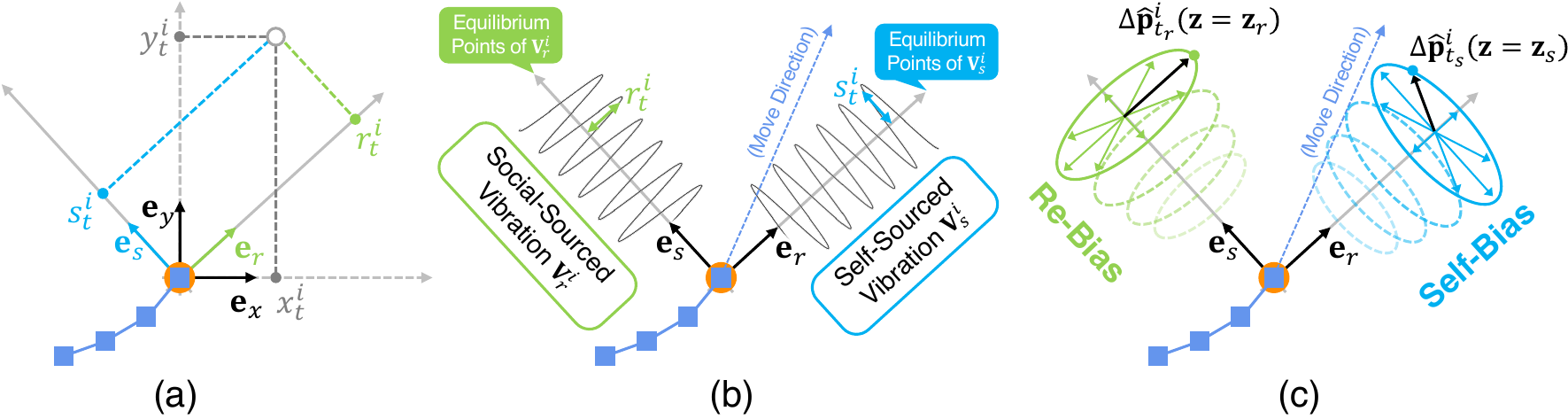}
    \caption{
        Descriptions of vibrations (b) and their sampled trajectory biases (c) in a coordinate transformation perspective.
    }
    \label{fig_biases_loss}
\end{figure}

\begin{figure*}[tbph]
    \centering
    \includegraphics[width=1.0\linewidth]{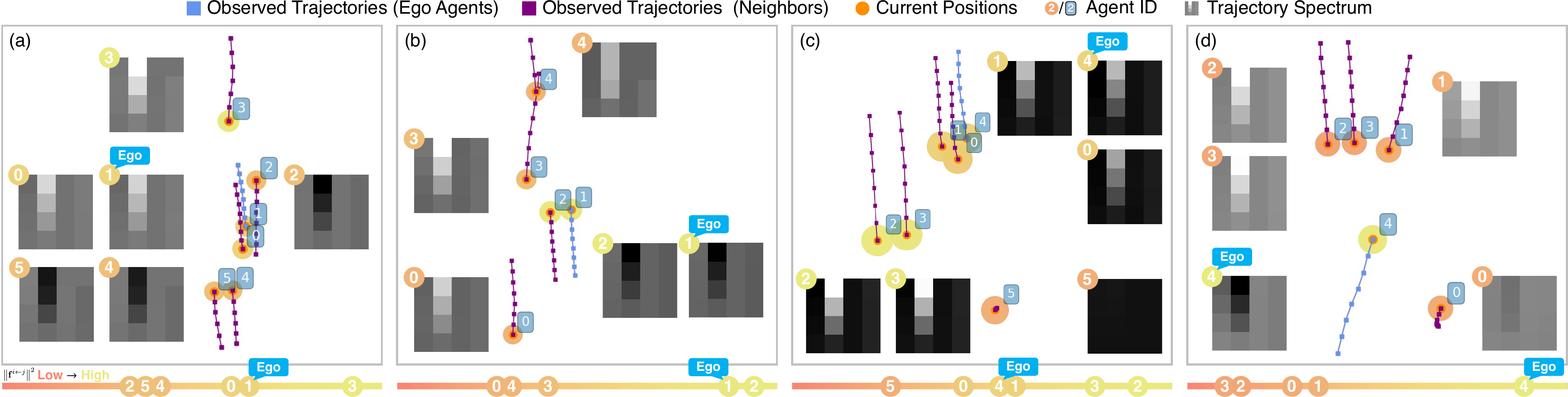}
    \caption{
        Visualized comparisons of spectrums (1D Haar transform, $\mathcal{T}(\mathbf{X}^j) \in \mathbb{R}^{4 \times 4}$ for trajectory $\mathbf{X}^j \in \mathbb{R}^{8 \times 2}$) and energy of resonance features with different neighbors to the same ego in ETH-UCY scenes.
        Lower energies are colored in {{\textcolor[HTML]{f9816f}{orange}}}, while higher ones in {{\textcolor[HTML]{e6e674}{green}}}.
    }
    \label{fig_resonance}
\end{figure*}


\begin{figure*}[tbph]
    \centering
    \includegraphics[width=1.0\linewidth]{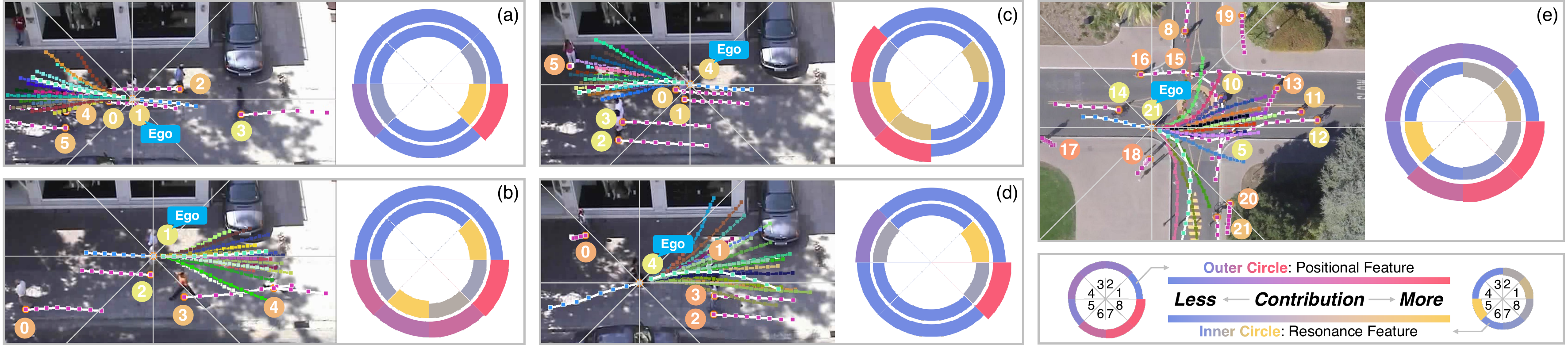}
    \caption{
        Visualized contribution comparisons of positional information $\mathbf{f}^{i \leftarrow j}_p$ (outer circles, high contribution ones are colored in {{\textcolor[HTML]{f8526c}{red}}}) and resonance feature $\mathbf{f}^{i \leftarrow j}$ (inner circles, high contributions in {{\textcolor[HTML]{f8c857}{yellow}}}) of neighbors (\EQUA{eq_fij}) after the angle-based resonance gathering.
    }
    \label{fig_resonance_pool}
\end{figure*}

\begin{figure}[tbph]
    \centering
    \includegraphics[width=0.73\linewidth]{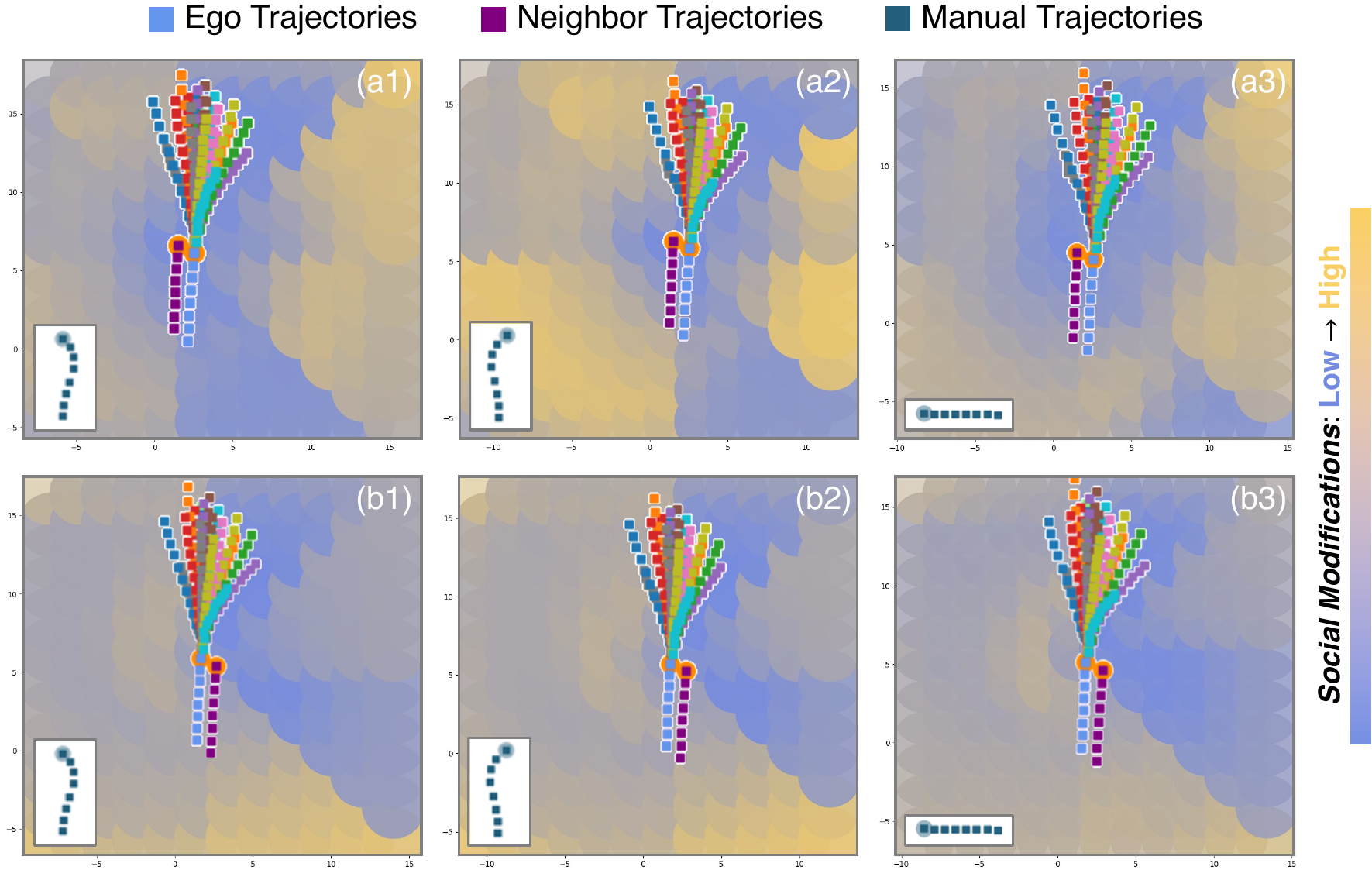}
    \caption{
        Spatial distribution $P(c(x, y | \mathbf{X}^m))$ of how much social modifications manual neighbor $m$ have made when forecasting.
    }
    \label{fig_traj}
\end{figure}

\NEWHALFLINE\textbf{Discussions on Vibrations.}
\FIG{fig_biases} visualizes several combinations of trajectory biases ($K=20$).
Here, we observe:

(i) Social-sourced vibrations (re-biases) present more \textbf{Nonlinearities}.
\MODELSHORT~uses two mirrored Transformers to predict two biases.
Their main difference is the target value for queries, better seen by comparing \EQUA{eq_selfbias,eq_rebias}.
In \FIG{fig_biases} (a1) to (a3), we see that each predicted self-bias is almost linearly distributed over time.
Even for the more challenging SDD ((a4) and (a5)) and NBA (a6) scenes, their nonlinearities are still limited, with the few non-linear ones mostly indicating slow or long-term changes.
In contrast, re-biases preserve distinguishing rapid and interactive changes.
Comparing \FIG{fig_biases} (a3), (b3), and (c3), it can be seen that a potential interactive behavior for the ego agent is to bypass the standingstill neighbor on the bottom-right side.
Also, several rapid changes have been posed in re-biases (\FIG{fig_biases} (b3)), like the {{\textcolor[HTML]{FF7410}{orange}}} one to the right, which appears to conduct a quick modification to achieve this goal.

(ii) Self-sourced and social-sourced vibrations \textbf{Vibrate alongside Different Directions}.
At any future step $t$, we observe that $\mathbf{V}^i_s(t,\mathbf{z})$ and $\mathbf{V}^i_r(t,\mathbf{z})$ distribute roughly around straight lines after sampling noise variable $\mathbf{z}$ for $K=20$ times.
We have drawn yellow dashed lines in \FIG{fig_biases} to better describe how they distribute\footnote{
    These dashed lines remain constant after training, but may vary during training or for different egos.
    See visualized discussions of such angular relations and how they change during training in \SUPP.
} when $t = t_h + t_f$.
According to these lines, it can be seen that the two vibrations grow in distinctly different directions, where self-sourced vibrations are better at capturing multipath (self-randomness) properties \emph{around} the movement direction, like the changing of orientations, while social-sourced vibrations vibrate \emph{alongside} this direction, aiming at randomizing their scheduled velocities to react to social behaviors.

(iii) Self-sourced and social-sourced vibrations \textbf{Vibrate Almost Vertically}.
Notably, each two of these dashed lines in \FIG{fig_biases} are almost vertical to each other (except for the NBA case).
This phenomenon can be explained as a decomposition of future trajectories and randomnesses alongside two learnable but almost mutually orthogonal directions $\mathbf{e}_s \bot \mathbf{e}_r$ (\FIG{fig_biases_loss} (a)).
For a predicted position $\hatbf{p}_t^i = (x^i_t, y^i_t)^\top := x^i_t \mathbf{e}_x + y^i_t \mathbf{e}_y$, it can be described as a coordinate transformation $(x^i_t, y^i_t) \to (s^i_t, r^i_t)$.
Here, $s^i_t$ and $r^i_t$ are the states (distances relative to the axis of equilibrium points, illustrated in \FIG{fig_biases_loss} (b)) of the two vibrations.
Intuitively, interactions between these vibrations may become minimized when they vibrate vertically, meaning that they tend to be independent of each other.
Under this circumstance, how the facing direction changes could be only attributed to the self-sourced vibration (forecasted as self-biases), while how the scheduled velocity changes to the other social-sourced one (re-biases) (\FIG{fig_biases_loss} (c)).
Such phenomena further validate our thought of treating trajectories as a superposition of decoupled vibrations from an explainable point of view.

\NEWHALFLINE\textbf{Discussions of Resonance Features.}
\MODELSHORT~simulates social interactions as the \emph{social resonance} inferred from currently observed vibrations (trajectories) of ego $i$ and all other neighbors.
We use the energy of resonance feature, \IE, $\|\mathbf{f}^{i \leftarrow j}\|^2$, to analyze how it distinguishes and compares vibrations of agent pair $\{i, j\}$.
While it is hard to tell if two resonance features are identical when their energies are the same, a larger energy difference may indicate a larger feature level difference
\footnote{
    Please see more feature-level discussions in \SUPP.
}.
\FIG{fig_resonance} illustrates trajectory spectrums and feature energies of different agent pairs.
We see that resonance features $\{\mathbf{f}^{i \leftarrow j}\}$ present different energies corresponding to different neighbors' vibrations (trajectories).
For example, neighbors with higher velocities may lead to higher energies and vice versa, especially for those standing still ones.
In particular, feature energies may differ over both ego's and neighbor's motion statuses, even if two agents share similar velocities.
Like shown in \FIG{fig_resonance} (a), the detailed spectral differences have been captured for ego $i = 1$ to distinguish neighbors $j \in \{0, 3\}$, although their trajectories are similar in the spatial-temporal space.

Notably, resonance features present a clear tendency to describe neighbors in distinct \emph{groups}.
For instance, for the ego agent $i = 1$ in \FIG{fig_resonance} (a), feature energies of resonance features between neighbors $j \in$ \{2, 4, 5\} are similar to each other but totally different from all others, and vice versa for groups \{0, 1\} or \{3\}.
This aligns with the scenario that pedestrians \{0, 1\} and \{4, 5\} do behave as two distinct groups.
The groupings are more pronounced in \FIG{fig_resonance} (c), where the walking-together agents \{0, 1, 4\} or \{2, 3\} share similar feature energies, while their inter-group energy differences are still evident.
Thus, our assumption that social interactions are associated with spectral properties of vibrations could be partially validated, and \MODELSHORT~have learned to tell these spectral differences without manual interventions.
This also aligns the \emph{social resonance}, described as \emph{people tend to hang out with those who are on a similar spectrum}.

Also, we use trainable weights in the first linear layer in $T_r$, denoted as
$\mathbf{w} = \left(\begin{array}{c;{1pt/1pt}c;{1pt/1pt}c;{1pt/1pt}c}
    \mathbf{w}_e &
    \mathbf{w}_{\mathcal{R}, r} &
    \mathbf{w}_{\mathcal{R}, p} &
    \mathbf{w}_z \\
\end{array}\right)
$ (corresponding to different terms in \EQUA{eq_rebias}), to further roughly compare the contribution between resonance features and positional information in the resonance matrix $\mathbf{F}^i_{\mathcal{R}}$.
The resonance contribution is computed as $\|\mathbf{w}_{\mathcal{R}, r}\mathbf{f}^{i \leftarrow j}\|^2$, and the positional contribution as $\|\mathbf{w}_{\mathcal{R}, p}\mathbf{f}^{i \leftarrow j}\|^2$.
\FIG{fig_resonance_pool} visualizes these contributions in every angle-based partition.
It can be seen that both spectral and positional status matter when making predictions. 
\MODELSHORT~may pay more attention to those neighbors who own greater vibration differences to the ego agent.
For example, neighbor 3 (who owns the most feature energy) contributes the most in \FIG{fig_resonance_pool} (a).
In \FIG{fig_resonance_pool} (e), however, neighbor 17 catches the most resonance attention, though its feature energy is way lower than the ego agent 21.
Notably, the 5th partition in \FIG{fig_resonance_pool} (e) has almost been paid the minimal attention when only considering positional information.
As a compromise, partition 5 no longer contributes the most finally.
This verifies the modeling capacity of resonance matrix $\mathbf{F}^i_{\mathcal{R}}$ in gathering both spectral and positional properties of all neighbors and their vibrations.

\NEWHALFLINE\textbf{Further Discussions of Social Resonance.}
We now discuss how predictions would be \emph{socially} modified by a single neighbor.
Considering a neighbor $m$ located at $\mathbf{p}^m_{t_h} = (x^m, y^m)^\top$, we define term $c(x^m, y^m | \mathbf{X}^m)$ to describe the absolute trajectory modification when using and not using this neighbor to forecast\footnote{
    See definitions and further descriptions in \SUPP.
}.
This is actually a causality validation \cite{granger1969investigating} that measures the effectiveness of social-sourced vibrations to handle social interactions, where no modification means causalities are not captured.
Given variable $\Delta \mathbf{p} = (\Delta x, \Delta y)^\top$, we first generate a series of translated trajectories $\{\mathbf{X}^m + \Delta \mathbf{p}\}$.
Denote $x = x^m + \Delta x$ and $y = y^m + \Delta y$, \FIG{fig_traj} visualizes the spatial distribution of modifications $c(x, y| \mathbf{X}^m)$ over different $x$-$y$ and $\mathbf{X}^m$ settings.
Here, we observe that egos present distinguishable social reactions under different situations.
For example, more modifications will be made when considering a left-turn neighbor to the right of the ego in \FIG{fig_traj} (a1).
Differently, a right-turn neighbor may catch more attention when positioned to the left.
Also, different modifications have made even though the trajectories of neighbors, also egos themselves, are kept constant.
Comparing \FIG{fig_traj} (a1) and (b1), the right agent cares more about others around its right side, while its left side has been taken over by its teammate.
Such results could verify that causalities within social behaviors could be captured by the social-sourced vibration, also the usefulness of the resonance-like modeling.

\section{Conclusion}

Inspired by vibrations and their resonance properties, we propose \MODEL~(\MODELSHORT) to forecast pedestrian trajectories as the superposition of vibrations.
It decomposes randomnesses in trajectories as multiple vibrations, thus learning different randomnesses to respond to different trajectory determinants.
It also uses a resonance-like manner to represent social interactions when forecasting.
Experiments have verified the usefulness of \MODELSHORT~quantitatively and qualitatively.


{
    \small
    \bibliographystyle{ieeenat_fullname}
    \bibliography{ref.bib}

\begin{thebibliography}{92}
\providecommand{\natexlab}[1]{#1}
\providecommand{\url}[1]{\texttt{#1}}
\expandafter\ifx\csname urlstyle\endcsname\relax
  \providecommand{\doi}[1]{doi: #1}\else
  \providecommand{\doi}{doi: \begingroup \urlstyle{rm}\Url}\fi

\bibitem[Alahi et~al.(2016)Alahi, Goel, Ramanathan, Robicquet, Fei-Fei, and
  Savarese]{alahi2016social}
Alexandre Alahi, Kratarth Goel, Vignesh Ramanathan, Alexandre Robicquet, Li
  Fei-Fei, and Silvio Savarese.
\newblock Social lstm: Human trajectory prediction in crowded spaces.
\newblock In \emph{Proceedings of the IEEE conference on computer vision and
  pattern recognition}, pages 961--971, 2016.

\bibitem[Alahi et~al.(2017)Alahi, Ramanathan, Goel, Robicquet, Sadeghian,
  Fei-Fei, and Savarese]{alahi2017learning}
Alexandre Alahi, Vignesh Ramanathan, Kratarth Goel, Alexandre Robicquet, Amir~A
  Sadeghian, Li Fei-Fei, and Silvio Savarese.
\newblock Learning to predict human behavior in crowded scenes.
\newblock In \emph{Group and Crowd Behavior for Computer Vision}, pages
  183--207. Elsevier, 2017.

\bibitem[Anvari et~al.(2015)Anvari, Bell, Sivakumar, and
  Ochieng]{anvari2015modelling}
Bani Anvari, Michael~GH Bell, Aruna Sivakumar, and Washington~Y Ochieng.
\newblock Modelling shared space users via rule-based social force model.
\newblock \emph{Transportation Research Part C: Emerging Technologies},
  51:\penalty0 83--103, 2015.

\bibitem[Bae et~al.(2024{\natexlab{a}})Bae, Lee, and Jeon]{bae2024can}
Inhwan Bae, Junoh Lee, and Hae-Gon Jeon.
\newblock Can language beat numerical regression? language-based multimodal
  trajectory prediction.
\newblock In \emph{Proceedings of the IEEE/CVF Conference on Computer Vision
  and Pattern Recognition}, pages 753--766, 2024{\natexlab{a}}.

\bibitem[Bae et~al.(2024{\natexlab{b}})Bae, Park, and
  Jeon]{bae2024singulartrajectory}
Inhwan Bae, Young-Jae Park, and Hae-Gon Jeon.
\newblock Singulartrajectory: Universal trajectory predictor using diffusion
  model.
\newblock \emph{arXiv preprint arXiv:2403.18452}, 2024{\natexlab{b}}.

\bibitem[Caesar et~al.(2019)Caesar, Bankiti, Lang, Vora, Liong, Xu, Krishnan,
  Pan, Baldan, and Beijbom]{caesar2019nuscenes}
Holger Caesar, Varun Bankiti, Alex~H. Lang, Sourabh Vora, Venice~Erin Liong,
  Qiang Xu, Anush Krishnan, Yu Pan, Giancarlo Baldan, and Oscar Beijbom.
\newblock nuscenes: A multimodal dataset for autonomous driving.
\newblock \emph{arXiv preprint arXiv:1903.11027}, 2019.

\bibitem[Caesar et~al.(2020)Caesar, Bankiti, Lang, Vora, Liong, Xu, Krishnan,
  Pan, Baldan, and Beijbom]{caesar2020nuscenes}
Holger Caesar, Varun Bankiti, Alex~H Lang, Sourabh Vora, Venice~Erin Liong,
  Qiang Xu, Anush Krishnan, Yu Pan, Giancarlo Baldan, and Oscar Beijbom.
\newblock nuscenes: A multimodal dataset for autonomous driving.
\newblock In \emph{Proceedings of the IEEE/CVF conference on computer vision
  and pattern recognition}, pages 11621--11631, 2020.

\bibitem[Chai et~al.(2019)Chai, Sapp, Bansal, and Anguelov]{chai2019multipath}
Yuning Chai, Benjamin Sapp, Mayank Bansal, and Dragomir Anguelov.
\newblock Multipath: Multiple probabilistic anchor trajectory hypotheses for
  behavior prediction.
\newblock \emph{arXiv preprint arXiv:1910.05449}, 2019.

\bibitem[Chen et~al.(2021)Chen, Li, Zhou, Ren, and Lu]{chen2021personalized}
Guangyi Chen, Junlong Li, Nuoxing Zhou, Liangliang Ren, and Jiwen Lu.
\newblock Personalized trajectory prediction via distribution discrimination.
\newblock In \emph{Proceedings of the IEEE/CVF International Conference on
  Computer Vision}, pages 15580--15589, 2021.

\bibitem[Chen et~al.(2022)Chen, Ivanovic, and Pavone]{chen2022scept}
Yuxiao Chen, Boris Ivanovic, and Marco Pavone.
\newblock Scept: Scene-consistent, policy-based trajectory predictions for
  planning.
\newblock In \emph{Proceedings of the IEEE/CVF Conference on Computer Vision
  and Pattern Recognition}, pages 17103--17112, 2022.

\bibitem[Chib and Singh(2024{\natexlab{a}})]{chib2024enhancing}
Pranav~Singh Chib and Pravendra Singh.
\newblock Enhancing trajectory prediction through self-supervised waypoint
  distortion prediction.
\newblock In \emph{International Conference on Machine Learning}, pages
  8403--8416. PMLR, 2024{\natexlab{a}}.

\bibitem[Chib and Singh(2024{\natexlab{b}})]{chib2024lg}
Pranav~Singh Chib and Pravendra Singh.
\newblock Lg-traj: Llm guided pedestrian trajectory prediction.
\newblock \emph{arXiv preprint arXiv:2403.08032}, 2024{\natexlab{b}}.

\bibitem[Chib et~al.(2024)Chib, Nath, Kabra, Gupta, and Singh]{chib2024ms}
Pranav~Singh Chib, Achintya Nath, Paritosh Kabra, Ishu Gupta, and Pravendra
  Singh.
\newblock Ms-tip: Imputation aware pedestrian trajectory prediction.
\newblock In \emph{International Conference on Machine Learning}, pages
  8389--8402. PMLR, 2024.

\bibitem[Choi et~al.(2024)Choi, Mercurius, Shabestary, and
  Rasouli]{choi2024dice}
Younwoo Choi, Ray~Coden Mercurius, Soheil Mohamad~Alizadeh Shabestary, and Amir
  Rasouli.
\newblock Dice: Diverse diffusion model with scoring for trajectory prediction.
\newblock In \emph{2024 IEEE Intelligent Vehicles Symposium (IV)}, pages
  3023--3029. IEEE, 2024.

\bibitem[Duan et~al.(2022)Duan, Wang, Long, Zhou, Zheng, Shi, and
  Hua]{duan2022complementary}
Jinghai Duan, Le Wang, Chengjiang Long, Sanping Zhou, Fang Zheng, Liushuai Shi,
  and Gang Hua.
\newblock Complementary attention gated network for pedestrian trajectory
  prediction.
\newblock In \emph{Proceedings of the AAAI Conference on Artificial
  Intelligence}, pages 542--550, 2022.

\bibitem[Fernando et~al.(2018)Fernando, Denman, Sridharan, and
  Fookes]{fernando2018soft}
Tharindu Fernando, Simon Denman, Sridha Sridharan, and Clinton Fookes.
\newblock Soft+ hardwired attention: An lstm framework for human trajectory
  prediction and abnormal event detection.
\newblock \emph{Neural networks}, 108:\penalty0 466--478, 2018.

\bibitem[Ge et~al.(2023)Ge, Song, and Huang]{ge2023causal}
Chunjiang Ge, Shiji Song, and Gao Huang.
\newblock Causal intervention for human trajectory prediction with cross
  attention mechanism.
\newblock In \emph{Proceedings of the AAAI Conference on Artificial
  Intelligence}, pages 658--666, 2023.

\bibitem[Granger(1969)]{granger1969investigating}
Clive~WJ Granger.
\newblock Investigating causal relations by econometric models and
  cross-spectral methods.
\newblock \emph{Econometrica: journal of the Econometric Society}, pages
  424--438, 1969.

\bibitem[Gupta et~al.(2018)Gupta, Johnson, Fei-Fei, Savarese, and
  Alahi]{gupta2018social}
Agrim Gupta, Justin Johnson, Li Fei-Fei, Silvio Savarese, and Alexandre Alahi.
\newblock Social gan: Socially acceptable trajectories with generative
  adversarial networks.
\newblock In \emph{Proceedings of the IEEE Conference on Computer Vision and
  Pattern Recognition}, pages 2255--2264, 2018.

\bibitem[Helbing and Molnar(1995)]{helbing1995social}
Dirk Helbing and Peter Molnar.
\newblock Social force model for pedestrian dynamics.
\newblock \emph{Physical review E}, 51\penalty0 (5):\penalty0 4282, 1995.

\bibitem[Huang et~al.(2019)Huang, Bi, Li, Mao, and Wang]{huang2019stgat}
Yingfan Huang, Huikun Bi, Zhaoxin Li, Tianlu Mao, and Zhaoqi Wang.
\newblock Stgat: Modeling spatial-temporal interactions for human trajectory
  prediction.
\newblock In \emph{Proceedings of the IEEE International Conference on Computer
  Vision}, pages 6272--6281, 2019.

\bibitem[Huang et~al.(2021)Huang, Wang, Pi, Song, and Yang]{huang2021lstm}
Zhi Huang, Jun Wang, Lei Pi, Xiaolin Song, and Lingfang Yang.
\newblock Lstm based trajectory prediction model for cyclist utilizing multiple
  interactions with environment.
\newblock \emph{Pattern Recognition}, 112:\penalty0 107800, 2021.

\bibitem[Jain et~al.(2016)Jain, Zamir, Savarese, and
  Saxena]{jain2016structural}
Ashesh Jain, Amir~R Zamir, Silvio Savarese, and Ashutosh Saxena.
\newblock Structural-rnn: Deep learning on spatio-temporal graphs.
\newblock In \emph{Proceedings of the ieee conference on computer vision and
  pattern recognition}, pages 5308--5317, 2016.

\bibitem[Karras et~al.(2019)Karras, Laine, and Aila]{karras2019style}
Tero Karras, Samuli Laine, and Timo Aila.
\newblock A style-based generator architecture for generative adversarial
  networks.
\newblock In \emph{Proceedings of the IEEE/CVF conference on computer vision
  and pattern recognition}, pages 4401--4410, 2019.

\bibitem[Kim et~al.(2017)Kim, Kang, Kim, Lee, Chung, and
  Choi]{kim2017probabilistic}
ByeoungDo Kim, Chang~Mook Kang, Jaekyum Kim, Seung~Hi Lee, Chung~Choo Chung,
  and Jun~Won Choi.
\newblock Probabilistic vehicle trajectory prediction over occupancy grid map
  via recurrent neural network.
\newblock In \emph{2017 IEEE 20th International Conference on Intelligent
  Transportation Systems (ITSC)}, pages 399--404. IEEE, 2017.

\bibitem[Kipf and Welling(2016)]{kipf2016semi}
Thomas~N Kipf and Max Welling.
\newblock Semi-supervised classification with graph convolutional networks.
\newblock \emph{arXiv preprint arXiv:1609.02907}, 2016.

\bibitem[Kosaraju et~al.(2019)Kosaraju, Sadeghian, Mart{\'\i}n-Mart{\'\i}n,
  Reid, Rezatofighi, and Savarese]{kosaraju2019social}
Vineet Kosaraju, Amir Sadeghian, Roberto Mart{\'\i}n-Mart{\'\i}n, Ian Reid,
  Hamid Rezatofighi, and Silvio Savarese.
\newblock Social-bigat: Multimodal trajectory forecasting using bicycle-gan and
  graph attention networks.
\newblock In \emph{Advances in Neural Information Processing Systems}, pages
  137--146, 2019.

\bibitem[Kothari and Alahi(2023)]{kothari2023safety}
Parth Kothari and Alexandre Alahi.
\newblock Safety-compliant generative adversarial networks for human trajectory
  forecasting.
\newblock \emph{IEEE Transactions on Intelligent Transportation Systems},
  24\penalty0 (4):\penalty0 4251--4261, 2023.

\bibitem[Lee et~al.(2022)Lee, Sohn, Moon, Yoon, Kapadia, and
  Pavlovic]{lee2022muse}
Mihee Lee, Samuel~S Sohn, Seonghyeon Moon, Sejong Yoon, Mubbasir Kapadia, and
  Vladimir Pavlovic.
\newblock Muse-vae: Multi-scale vae for environment-aware long term trajectory
  prediction.
\newblock In \emph{Proceedings of the IEEE/CVF Conference on Computer Vision
  and Pattern Recognition}, pages 2221--2230, 2022.

\bibitem[Lee et~al.(2017)Lee, Choi, Vernaza, Choy, Torr, and
  Chandraker]{lee2017desire}
Namhoon Lee, Wongun Choi, Paul Vernaza, Christopher~B Choy, Philip~HS Torr, and
  Manmohan Chandraker.
\newblock Desire: Distant future prediction in dynamic scenes with interacting
  agents.
\newblock In \emph{Proceedings of the IEEE Conference on Computer Vision and
  Pattern Recognition}, pages 336--345, 2017.

\bibitem[Lerner et~al.(2007)Lerner, Chrysanthou, and
  Lischinski]{lerner2007crowds}
Alon Lerner, Yiorgos Chrysanthou, and Dani Lischinski.
\newblock Crowds by example.
\newblock \emph{Computer Graphics Forum}, 26\penalty0 (3):\penalty0 655--664,
  2007.

\bibitem[Li et~al.(2021{\natexlab{a}})Li, Yang, Ma, Malla, Tomizuka, and
  Choi]{li2021rain}
Jiachen Li, Fan Yang, Hengbo Ma, Srikanth Malla, Masayoshi Tomizuka, and Chiho
  Choi.
\newblock Rain: Reinforced hybrid attention inference network for motion
  forecasting.
\newblock In \emph{Proceedings of the IEEE/CVF International Conference on
  Computer Vision}, pages 16096--16106, 2021{\natexlab{a}}.

\bibitem[Li et~al.(2022)Li, Pagnucco, and Song]{li2022graph}
Lihuan Li, Maurice Pagnucco, and Yang Song.
\newblock Graph-based spatial transformer with memory replay for multi-future
  pedestrian trajectory prediction.
\newblock In \emph{Proceedings of the IEEE/CVF Conference on Computer Vision
  and Pattern Recognition}, pages 2231--2241, 2022.

\bibitem[Li et~al.(2024)Li, Li, Ren, Chen, Yuan, and Wang]{li2024bcdiff}
Rongqing Li, Changsheng Li, Dongchun Ren, Guangyi Chen, Ye Yuan, and Guoren
  Wang.
\newblock Bcdiff: Bidirectional consistent diffusion for instantaneous
  trajectory prediction.
\newblock \emph{Advances in Neural Information Processing Systems}, 36, 2024.

\bibitem[Li et~al.(2021{\natexlab{b}})Li, Zhou, Yi, and Gall]{li2021spatial}
Shijie Li, Yanying Zhou, Jinhui Yi, and Juergen Gall.
\newblock Spatial-temporal consistency network for low-latency trajectory
  forecasting.
\newblock In \emph{Proceedings of the IEEE/CVF International Conference on
  Computer Vision (ICCV)}, pages 1940--1949, 2021{\natexlab{b}}.

\bibitem[Liang et~al.(2019)Liang, Jiang, Niebles, Hauptmann, and
  Fei-Fei]{liang2019peeking}
Junwei Liang, Lu Jiang, Juan~Carlos Niebles, Alexander~G Hauptmann, and Li
  Fei-Fei.
\newblock Peeking into the future: Predicting future person activities and
  locations in videos.
\newblock In \emph{Proceedings of the IEEE Conference on Computer Vision and
  Pattern Recognition}, pages 5725--5734, 2019.

\bibitem[Liang et~al.(2020)Liang, Jiang, and Hauptmann]{liang2020simaug}
Junwei Liang, Lu Jiang, and Alexander Hauptmann.
\newblock Simaug: Learning robust representations from simulation for
  trajectory prediction.
\newblock In \emph{Proceedings of the European conference on computer vision
  (ECCV)}, 2020.

\bibitem[Lin et~al.(2024)Lin, Liang, Lai, and Hu]{lin2024progressive}
Xiaotong Lin, Tianming Liang, Jianhuang Lai, and Jian-Fang Hu.
\newblock Progressive pretext task learning for human trajectory prediction.
\newblock In \emph{European Conference on Computer Vision}, pages 197--214.
  Springer, 2024.

\bibitem[Linou et~al.(2016)Linou, Linou, and de~Boer]{linou2016nba}
Kostya Linou, Dzmitryi Linou, and Martijn de Boer.
\newblock Nba player movements.
\newblock https://github.com/linouk23/NBA-Player-Movements, 2016.

\bibitem[Liu et~al.(2021)Liu, Chen, Liu, and Shi]{liu2021avgcn}
Congcong Liu, Yuying Chen, Ming Liu, and Bertram~E Shi.
\newblock Avgcn: Trajectory prediction using graph convolutional networks
  guided by human attention.
\newblock In \emph{2021 IEEE International Conference on Robotics and
  Automation (ICRA)}, pages 14234--14240. IEEE, 2021.

\bibitem[Liu et~al.(2024)Liu, Ye, Wang, Li, Sheng, and Yao]{liu2024uncertainty}
Yao Liu, Zesheng Ye, Rui Wang, Binghao Li, Quan~Z Sheng, and Lina Yao.
\newblock Uncertainty-aware pedestrian trajectory prediction via distributional
  diffusion.
\newblock \emph{Knowledge-Based Systems}, page 111862, 2024.

\bibitem[Luber et~al.(2010)Luber, Stork, Tipaldi, and Arras]{luber2010people}
Matthias Luber, Johannes~A Stork, Gian~Diego Tipaldi, and Kai~O Arras.
\newblock People tracking with human motion predictions from social forces.
\newblock In \emph{2010 IEEE international conference on robotics and
  automation}, pages 464--469. IEEE, 2010.

\bibitem[Lv and Yuan(2023)]{lv2023skgacn}
Kai Lv and Liang Yuan.
\newblock Skgacn: social knowledge-guided graph attention convolutional network
  for human trajectory prediction.
\newblock \emph{IEEE Transactions on Instrumentation and Measurement}, 2023.

\bibitem[Ma et~al.(2019)Ma, Zhu, Zhang, Yang, Wang, and
  Manocha]{ma2019trafficpredict}
Yuexin Ma, Xinge Zhu, Sibo Zhang, Ruigang Yang, Wenping Wang, and Dinesh
  Manocha.
\newblock Trafficpredict: Trajectory prediction for heterogeneous
  traffic-agents.
\newblock In \emph{Proceedings of the AAAI Conference on Artificial
  Intelligence}, pages 6120--6127, 2019.

\bibitem[Maeda and Ukita(2023)]{maeda2023fast}
Takahiro Maeda and Norimichi Ukita.
\newblock Fast inference and update of probabilistic density estimation on
  trajectory prediction.
\newblock In \emph{Proceedings of the IEEE/CVF International Conference on
  Computer Vision}, pages 9795--9805, 2023.

\bibitem[Makansi et~al.(2021)Makansi, Von~K{\"u}gelgen, Locatello, Gehler,
  Janzing, Brox, and Sch{\"o}lkopf]{makansi2021you}
Osama Makansi, Julius Von~K{\"u}gelgen, Francesco Locatello, Peter Gehler,
  Dominik Janzing, Thomas Brox, and Bernhard Sch{\"o}lkopf.
\newblock You mostly walk alone: Analyzing feature attribution in trajectory
  prediction.
\newblock \emph{arXiv preprint arXiv:2110.05304}, 2021.

\bibitem[Mangalam et~al.(2020)Mangalam, Girase, Agarwal, Lee, Adeli, Malik, and
  Gaidon]{mangalam2020not}
Karttikeya Mangalam, Harshayu Girase, Shreyas Agarwal, Kuan-Hui Lee, Ehsan
  Adeli, Jitendra Malik, and Adrien Gaidon.
\newblock It is not the journey but the destination: Endpoint conditioned
  trajectory prediction.
\newblock In \emph{European Conference on Computer Vision}, pages 759--776,
  2020.

\bibitem[Mangalam et~al.(2021)Mangalam, An, Girase, and Malik]{mangalam2020s}
Karttikeya Mangalam, Yang An, Harshayu Girase, and Jitendra Malik.
\newblock From goals, waypoints \& paths to long term human trajectory
  forecasting.
\newblock In \emph{Proceedings of the IEEE/CVF International Conference on
  Computer Vision}, pages 15233--15242, 2021.

\bibitem[Mao et~al.(2020)Mao, Liu, and Salzmann]{mao2020history}
Wei Mao, Miaomiao Liu, and Mathieu Salzmann.
\newblock History repeats itself: Human motion prediction via motion attention.
\newblock In \emph{European Conference on Computer Vision}, pages 474--489.
  Springer, 2020.

\bibitem[Mao et~al.(2023)Mao, Xu, Zhu, Chen, and Wang]{mao2023leapfrog}
Weibo Mao, Chenxin Xu, Qi Zhu, Siheng Chen, and Yanfeng Wang.
\newblock Leapfrog diffusion model for stochastic trajectory prediction.
\newblock In \emph{Proceedings of the IEEE/CVF Conference on Computer Vision
  and Pattern Recognition}, pages 5517--5526, 2023.

\bibitem[Marchetti et~al.(2024)Marchetti, Becattini, Seidenari, and
  Del~Bimbo]{marchetti2024smemo}
Francesco Marchetti, Federico Becattini, Lorenzo Seidenari, and Alberto
  Del~Bimbo.
\newblock Smemo: social memory for trajectory forecasting.
\newblock \emph{IEEE Transactions on Pattern Analysis and Machine
  Intelligence}, 2024.

\bibitem[Mehran et~al.(2009)Mehran, Oyama, and Shah]{mehran2009abnormal}
Ramin Mehran, Alexis Oyama, and Mubarak Shah.
\newblock Abnormal crowd behavior detection using social force model.
\newblock In \emph{2009 IEEE Conference on Computer Vision and Pattern
  Recognition}, pages 935--942. IEEE, 2009.

\bibitem[Mohamed et~al.(2020)Mohamed, Qian, Elhoseiny, and
  Claudel]{mohamed2020social}
Abduallah Mohamed, Kun Qian, Mohamed Elhoseiny, and Christian Claudel.
\newblock Social-stgcnn: A social spatio-temporal graph convolutional neural
  network for human trajectory prediction.
\newblock In \emph{Proceedings of the IEEE/CVF Conference on Computer Vision
  and Pattern Recognition}, pages 14424--14432, 2020.

\bibitem[Neumeier et~al.(2021)Neumeier, Botsch, Tollk{\"u}hn, and
  Berberich]{neumeier2021variational}
Marion Neumeier, Michael Botsch, Andreas Tollk{\"u}hn, and Thomas Berberich.
\newblock Variational autoencoder-based vehicle trajectory prediction with an
  interpretable latent space.
\newblock In \emph{2021 IEEE International Intelligent Transportation Systems
  Conference (ITSC)}, pages 820--827. IEEE, 2021.

\bibitem[Park et~al.(2024)Park, Jeong, Yoon, Jeong, and Yoon]{park2024t4p}
Daehee Park, Jaeseok Jeong, Sung-Hoon Yoon, Jaewoo Jeong, and Kuk-Jin Yoon.
\newblock T4p: Test-time training of trajectory prediction via masked
  autoencoder and actor-specific token memory.
\newblock In \emph{Proceedings of the IEEE/CVF Conference on Computer Vision
  and Pattern Recognition}, pages 15065--15076, 2024.

\bibitem[Pellegrini et~al.(2009)Pellegrini, Ess, Schindler, and
  Van~Gool]{pellegrini2009youll}
Stefano Pellegrini, Andreas Ess, Konrad Schindler, and Luc Van~Gool.
\newblock You'll never walk alone: Modeling social behavior for multi-target
  tracking.
\newblock In \emph{2009 IEEE 12th International Conference on Computer Vision},
  pages 261--268. IEEE, 2009.

\bibitem[Phong et~al.(2024)Phong, Wu, Yu, Cai, Zheng, and Hsu]{phong2024truly}
Tran Phong, Haoran Wu, Cunjun Yu, Panpan Cai, Sifa Zheng, and David Hsu.
\newblock What truly matters in trajectory prediction for autonomous driving?
\newblock \emph{Advances in Neural Information Processing Systems}, 36, 2024.

\bibitem[Quan et~al.(2021)Quan, Zhu, Wu, and Yang]{quan2021holistic}
Ruijie Quan, Linchao Zhu, Yu Wu, and Yi Yang.
\newblock Holistic lstm for pedestrian trajectory prediction.
\newblock \emph{IEEE transactions on image processing}, 30:\penalty0
  3229--3239, 2021.

\bibitem[Rehder and Kloeden(2015)]{rehder2015goal}
Eike Rehder and Horst Kloeden.
\newblock Goal-directed pedestrian prediction.
\newblock In \emph{Proceedings of the IEEE International Conference on Computer
  Vision Workshops}, pages 50--58, 2015.

\bibitem[Rehder et~al.(2018)Rehder, Wirth, Lauer, and
  Stiller]{rehder2018pedestrian}
Eike Rehder, Florian Wirth, Martin Lauer, and Christoph Stiller.
\newblock Pedestrian prediction by planning using deep neural networks.
\newblock In \emph{2018 IEEE International Conference on Robotics and
  Automation (ICRA)}, pages 5903--5908. IEEE, 2018.

\bibitem[Robicquet et~al.(2016)Robicquet, Sadeghian, Alahi, and
  Savarese]{robicquet2016learning}
Alexandre Robicquet, Amir Sadeghian, Alexandre Alahi, and Silvio Savarese.
\newblock Learning social etiquette: Human trajectory understanding in crowded
  scenes.
\newblock In \emph{European conference on computer vision}, pages 549--565.
  Springer, 2016.

\bibitem[Rossi et~al.(2021)Rossi, Paolanti, Pierdicca, and
  Frontoni]{rossi2021human}
Luca Rossi, Marina Paolanti, Roberto Pierdicca, and Emanuele Frontoni.
\newblock Human trajectory prediction and generation using lstm models and
  gans.
\newblock \emph{Pattern Recognition}, 120:\penalty0 108136, 2021.

\bibitem[Saadatnejad et~al.(2022)Saadatnejad, Ju, and
  Alahi]{saadatnejad2022pedestrian}
Saeed Saadatnejad, Yi~Zhou Ju, and Alexandre Alahi.
\newblock Pedestrian 3d bounding box prediction.
\newblock \emph{arXiv preprint arXiv:2206.14195}, 2022.

\bibitem[Sadeghian et~al.(2019)Sadeghian, Kosaraju, Sadeghian, Hirose,
  Rezatofighi, and Savarese]{sadeghian2019sophie}
Amir Sadeghian, Vineet Kosaraju, Ali Sadeghian, Noriaki Hirose, Hamid
  Rezatofighi, and Silvio Savarese.
\newblock Sophie: An attentive gan for predicting paths compliant to social and
  physical constraints.
\newblock In \emph{Proceedings of the IEEE Conference on Computer Vision and
  Pattern Recognition}, pages 1349--1358, 2019.

\bibitem[Saleh et~al.(2020)Saleh, Aliakbarian, Salzmann, and
  Gould]{saleh2020artist}
Fatemeh Saleh, Sadegh Aliakbarian, Mathieu Salzmann, and Stephen Gould.
\newblock Artist: Autoregressive trajectory inpainting and scoring for
  tracking.
\newblock \emph{arXiv preprint arXiv:2004.07482}, 2020.

\bibitem[Salzmann et~al.(2020)Salzmann, Ivanovic, Chakravarty, and
  Pavone]{salzmann2020trajectron}
Tim Salzmann, Boris Ivanovic, Punarjay Chakravarty, and Marco Pavone.
\newblock Trajectron++: Dynamically-feasible trajectory forecasting with
  heterogeneous data.
\newblock In \emph{Proceedings of the European conference on computer vision
  (ECCV)}, pages 683--700. Springer, 2020.

\bibitem[Shafiee et~al.(2021)Shafiee, Padir, and
  Elhamifar]{shafiee2021introvert}
Nasim Shafiee, Taskin Padir, and Ehsan Elhamifar.
\newblock Introvert: Human trajectory prediction via conditional 3d attention.
\newblock In \emph{Proceedings of the IEEE/CVF Conference on Computer Vision
  and Pattern Recognition}, pages 16815--16825, 2021.

\bibitem[Shi et~al.(2023)Shi, Wang, Long, Zhou, Tang, Zheng, and
  Hua]{shi2023representing}
Liushuai Shi, Le Wang, Chengjiang Long, Sanping Zhou, Wei Tang, Nanning Zheng,
  and Gang Hua.
\newblock Representing multimodal behaviors with mean location for pedestrian
  trajectory prediction.
\newblock \emph{IEEE Transactions on Pattern Analysis and Machine
  Intelligence}, 2023.

\bibitem[Song et~al.(2021)Song, Chen, Li, Sun, Hou, Cui, Zhang, Xiong, and
  Wang]{song2020pedestrian}
Xiao Song, Kai Chen, Xu Li, Jinghan Sun, Baocun Hou, Yong Cui, Baochang Zhang,
  Gang Xiong, and Zilie Wang.
\newblock Pedestrian trajectory prediction based on deep convolutional lstm
  network.
\newblock \emph{IEEE Transactions on Intelligent Transportation Systems},
  22\penalty0 (6):\penalty0 3285--3302, 2021.

\bibitem[Su et~al.(2024)Su, Li, Wang, Zhou, and Li]{su2024unified}
Yuchao Su, Yuanman Li, Wei Wang, Jiantao Zhou, and Xia Li.
\newblock A unified environmental network for pedestrian trajectory prediction.
\newblock In \emph{Proceedings of the AAAI Conference on Artificial
  Intelligence}, pages 4970--4978, 2024.

\bibitem[Tran et~al.(2021)Tran, Le, and Tran]{tran2021goal}
Hung Tran, Vuong Le, and Truyen Tran.
\newblock Goal-driven long-term trajectory prediction.
\newblock In \emph{Proceedings of the IEEE/CVF Winter Conference on
  Applications of Computer Vision}, pages 796--805, 2021.

\bibitem[Vaswani et~al.(2017)Vaswani, Shazeer, Parmar, Uszkoreit, Jones, Gomez,
  Kaiser, and Polosukhin]{vaswani2017attention}
Ashish Vaswani, Noam Shazeer, Niki Parmar, Jakob Uszkoreit, Llion Jones,
  Aidan~N Gomez, {\L}ukasz Kaiser, and Illia Polosukhin.
\newblock Attention is all you need.
\newblock In \emph{Advances in neural information processing systems}, pages
  5998--6008, 2017.

\bibitem[Vemula et~al.(2018)Vemula, Muelling, and Oh]{vemula2018social}
Anirudh Vemula, Katharina Muelling, and Jean Oh.
\newblock Social attention: Modeling attention in human crowds.
\newblock In \emph{2018 IEEE international Conference on Robotics and
  Automation (ICRA)}, pages 1--7. IEEE, 2018.

\bibitem[Wang et~al.(2022)Wang, Ye, Gu, and Chen]{wang2022ltp}
Jingke Wang, Tengju Ye, Ziqing Gu, and Junbo Chen.
\newblock Ltp: Lane-based trajectory prediction for autonomous driving.
\newblock In \emph{Proceedings of the IEEE/CVF Conference on Computer Vision
  and Pattern Recognition}, pages 17134--17142, 2022.

\bibitem[Wong et~al.(2022)Wong, Xia, Hong, Peng, Yuan, Cao, Yang, and
  You]{wong2022view}
Conghao Wong, Beihao Xia, Ziming Hong, Qinmu Peng, Wei Yuan, Qiong Cao, Yibo
  Yang, and Xinge You.
\newblock View vertically: A hierarchical network for trajectory prediction via
  fourier spectrums.
\newblock In \emph{European Conference on Computer Vision}, pages 682--700.
  Springer, 2022.

\bibitem[Wong et~al.(2023)Wong, Xia, Peng, Yuan, and You]{wong2021msn}
Conghao Wong, Beihao Xia, Qinmu Peng, Wei Yuan, and Xinge You.
\newblock Msn: multi-style network for trajectory prediction.
\newblock \emph{IEEE Transactions on Intelligent Transportation Systems},
  24:\penalty0 9751 -- 9766, 2023.

\bibitem[Wong et~al.(2024{\natexlab{a}})Wong, Xia, Zou, Wang, and
  You]{wong2023socialcircle}
Conghao Wong, Beihao Xia, Ziqian Zou, Yulong Wang, and Xinge You.
\newblock Socialcircle: Learning the angle-based social interaction
  representation for pedestrian trajectory prediction.
\newblock In \emph{Proceedings of the IEEE/CVF Conference on Computer Vision
  and Pattern Recognition}, pages 19005--19015, 2024{\natexlab{a}}.

\bibitem[Wong et~al.(2024{\natexlab{b}})Wong, Xia, Zou, and
  You]{wong2024socialcircle+}
Conghao Wong, Beihao Xia, Ziqian Zou, and Xinge You.
\newblock Socialcircle+: Learning the angle-based conditioned interaction
  representation for pedestrian trajectory prediction.
\newblock \emph{arXiv preprint arXiv:2409.14984}, 2024{\natexlab{b}}.

\bibitem[Xia et~al.(2023)Xia, Wong, Xu, Peng, and You]{wong2023another}
Beihao Xia, Conghao Wong, Duanquan Xu, Qinmu Peng, and Xinge You.
\newblock Another vertical view: A hierarchical network for heterogeneous
  trajectory prediction via spectrums.
\newblock \emph{arXiv preprint arXiv:2304.05106}, 2023.

\bibitem[Xie et~al.(2024)Xie, Zhang, Xia, Xiao, Jiang, Zhou, Qin, and
  Chen]{xie2024pedestrian}
Jiajia Xie, Sheng Zhang, Beihao Xia, Zhu Xiao, Hongbo Jiang, Siwang Zhou, Zheng
  Qin, and Hongyang Chen.
\newblock Pedestrian trajectory prediction based on social interactions
  learning with random weights.
\newblock \emph{IEEE Transactions on Multimedia}, 2024.

\bibitem[Xu et~al.(2022{\natexlab{a}})Xu, Li, Ni, Zhang, and
  Chen]{xu2022groupnet}
Chenxin Xu, Maosen Li, Zhenyang Ni, Ya Zhang, and Siheng Chen.
\newblock Groupnet: Multiscale hypergraph neural networks for trajectory
  prediction with relational reasoning.
\newblock In \emph{Proceedings of the IEEE/CVF Conference on Computer Vision
  and Pattern Recognition (CVPR)}, pages 6498--6507, 2022{\natexlab{a}}.

\bibitem[Xu et~al.(2022{\natexlab{b}})Xu, Mao, Zhang, and Chen]{xu2022remember}
Chenxin Xu, Weibo Mao, Wenjun Zhang, and Siheng Chen.
\newblock Remember intentions: Retrospective-memory-based trajectory
  prediction.
\newblock In \emph{Proceedings of the IEEE/CVF Conference on Computer Vision
  and Pattern Recognition (CVPR)}, pages 6488--6497, 2022{\natexlab{b}}.

\bibitem[Xu et~al.(2023)Xu, Tan, Tan, Chen, Wang, Wang, and
  Wang]{xu2023eqmotion}
Chenxin Xu, Robby~T Tan, Yuhong Tan, Siheng Chen, Yu~Guang Wang, Xinchao Wang,
  and Yanfeng Wang.
\newblock Eqmotion: Equivariant multi-agent motion prediction with invariant
  interaction reasoning.
\newblock In \emph{Proceedings of the IEEE/CVF Conference on Computer Vision
  and Pattern Recognition}, pages 1410--1420, 2023.

\bibitem[Xu et~al.(2022{\natexlab{c}})Xu, Hayet, and
  Karamouzas]{xu2022socialvae}
Pei Xu, Jean-Bernard Hayet, and Ioannis Karamouzas.
\newblock Socialvae: Human trajectory prediction using timewise latents.
\newblock In \emph{European Conference on Computer Vision}, pages 511--528,
  2022{\natexlab{c}}.

\bibitem[Xu and Fu(2024)]{xu2024adapting}
Yi Xu and Yun Fu.
\newblock Adapting to length shift: Flexilength network for trajectory
  prediction.
\newblock In \emph{Proceedings of the IEEE/CVF Conference on Computer Vision
  and Pattern Recognition}, pages 15226--15237, 2024.

\bibitem[Yang et~al.(2025)Yang, Tian, Tian, Yu, Lu, Deng, and
  Sun]{yang2025sopermodel}
Heming Yang, Yu Tian, Changyuan Tian, Hongfeng Yu, Wanxuan Lu, Chubo Deng, and
  Xian Sun.
\newblock Sopermodel: Leveraging social perception for multi-agent trajectory
  prediction.
\newblock \emph{IEEE Transactions on Geoscience and Remote Sensing}, 2025.

\bibitem[Yu et~al.(2020)Yu, Ma, Ren, Zhao, and Yi]{yu2020spatio}
Cunjun Yu, Xiao Ma, Jiawei Ren, Haiyu Zhao, and Shuai Yi.
\newblock Spatio-temporal graph transformer networks for pedestrian trajectory
  prediction.
\newblock In \emph{European Conference on Computer Vision}, pages 507--523.
  Springer, 2020.

\bibitem[Yuan et~al.(2021)Yuan, Weng, Ou, and Kitani]{yuan2021agentformer}
Ye Yuan, Xinshuo Weng, Yanglan Ou, and Kris~M. Kitani.
\newblock Agentformer: Agent-aware transformers for socio-temporal multi-agent
  forecasting.
\newblock In \emph{Proceedings of the IEEE/CVF International Conference on
  Computer Vision (ICCV)}, pages 9813--9823, 2021.

\bibitem[Yue et~al.(2022)Yue, Manocha, and Wang]{yue2022human}
Jiangbei Yue, Dinesh Manocha, and He Wang.
\newblock Human trajectory prediction via neural social physics.
\newblock In \emph{European Conference on Computer Vision}, pages 376--394.
  Springer, 2022.

\bibitem[Zhang et~al.(2019)Zhang, Ouyang, Zhang, Xue, and Zheng]{zhang2019sr}
Pu Zhang, Wanli Ouyang, Pengfei Zhang, Jianru Xue, and Nanning Zheng.
\newblock Sr-lstm: State refinement for lstm towards pedestrian trajectory
  prediction.
\newblock In \emph{Proceedings of the IEEE Conference on Computer Vision and
  Pattern Recognition}, pages 12085--12094, 2019.

\bibitem[Zhang et~al.(2022)Zhang, Xue, Zhang, Zheng, and
  Ouyang]{zhang2020social}
Pu Zhang, Jianru Xue, Pengfei Zhang, Nanning Zheng, and Wanli Ouyang.
\newblock Social-aware pedestrian trajectory prediction via states refinement
  lstm.
\newblock \emph{IEEE transactions on pattern analysis and machine
  intelligence}, 44\penalty0 (5):\penalty0 2742--2759, 2022.

\bibitem[Zhou et~al.(2022)Zhou, Ye, Wang, Wu, and Lu]{zhou2022hivt}
Zikang Zhou, Luyao Ye, Jianping Wang, Kui Wu, and Kejie Lu.
\newblock Hivt: Hierarchical vector transformer for multi-agent motion
  prediction.
\newblock In \emph{Proceedings of the IEEE/CVF Conference on Computer Vision
  and Pattern Recognition}, pages 8823--8833, 2022.

\end{thebibliography}
}


\clearpage
\section*{\appendixname}
\appendix


\section{Metrics and Other Technical Details}

We use the best Average/Final Displacement Error over $K$ trajectories (minADE$_K$/minFDE$_K$) to measure performance \cite{alahi2016social,gupta2018social}.
In our main manuscript, we abbreviate them as ``ADE'' and ``FDE'' to save space.
For an ego agent $i$, denote the $k$th model output as $\hat{\mathbf{Y}}^i_k$, and the predicted position on the $t$th future frame as $\hat{\mathbf{p}_k}^i_t$, we have
\begin{align}
    {\mathrm{minADE}}_{K}(i)
    &= \min_{1 \leq k \leq K} \frac{1}{t_f}
    \sum_{t = t_h + 1}^{t_h + t_f}
    \left\Vert
        \mathbf{p}^i_t - 
        \hat{\mathbf{p}_k}^i_t
    \right\Vert_2, \\
    {\mathrm{minFDE}}_{K}(i)
    &= \min_{1 \leq k \leq K}
    \left\Vert
        \mathbf{p}^i_{t_h + t_f} - 
        \hat{\mathbf{p}_k}^i_{t_h + t_f}
    \right\Vert_2.
\end{align}

In addition, due to page limitations, some technical details are not presented in the main manuscript.
These parts are unrelated to our contributions but may be used to reproduce our work.
The linear trajectory (including both the linear fit and the linear base) will be translated by adding a constant vector to ensure that it can intersect with the observed trajectory at the current observation moment ($t = t_h$) to maintain continuity.
When computing the differential feature in the original Eq. (5), the outer product is also used to enhance the modeling capacities of trajectory spectrums in embedding networks $N_e$ and $N_{e, l}$ \cite{wong2023another}.
Also, following previous works \cite{ma2019trafficpredict}, category labels of agents have been encoded through one simple linear layer when embedding their trajectories only in nuScenes to learn the significant interclass differences of heterogeneous agents (vehicles).
In the original Eq. (8), we use linear-speed interpolation to obtain the final forecasted self-biases since it could be difficult for either pedestrians or vehicles to make sudden changes in their velocities (amplitude and direction) while in motion.
This means that when computing, we add the condition of equal velocities to the left and to the right of the interpolation keypoints.
Please refer to our code to learn how these details are implemented.

\section{Model Efficiency Analyses}

\begin{table}[tbp]
    \centering
    \footnotesize
    \begin{tabular}{lccccc}
        \toprule
        Method & Performance (SDD) & $t_{\mathrm{1}}$ & $t_{\mathrm{1k}}$ & Parameter \\

        \midrule
        \EVMODEL~\EVCITE & 6.57/10.49 & \C{28} & \C{112} & \C{1,976,864} \\
        \SCMODEL~\SCCITE & 6.54/10.36 & \C{34} & \C{119} & \C{1,989,536} \\
        \SCPMODEL~\SCPCITE & 6.44/10.22 &  41 & 126 & 1,990,177 \\

        \MODELSHORT~(variation a2) & 6.93/10.26 & \C{29} & \C{91} & 2,046,428 \\
        \MODELSHORT~(variation a8) & 6.33/10.45 & \C{34} & 211 & \C{1,242,924} \\
        \MODELSHORT~(full) & 6.27/10.02 & 56 & 263 & 3,149,192 \\

        \bottomrule
    \end{tabular}
    \caption{
        Comparisons of the average inference times (on an Apple M1 Mac mini (2020, 8GB Memory)) under the batch size of 1 and 1000 (denoted as $t_{\mathrm{1}}$ and $t_{\mathrm{1k}}$ correspondingly, reported in milliseconds), and parameter amount.
    }
    \label{tab_efficiency}
  \end{table}

We report the inference times and parameter amount of the proposed \MODELSHORT~in comparison to several newly published approaches in \TABLE{tab_efficiency}.
Due to the \emph{vibration-like} prediction strategy, two trajectory biases should be forecasted (through two mirrored Transformer backbones) when making the final prediction.
This means that the time-space efficiency of \MODELSHORT~is naturally a bit disadvantaged compared to other methods, requiring roughly twice as much parameterization and inference time as other methods.

By comparing several newly published methods that obtained similar prediction performance, it can be seen that \MODELSHORT~still has considerable time-space efficiency.
Please note that these efficiency experiments are conducted on one Apple M1 Mac mini (2020, 8GB Memory), whose computing performance is roughly the same as current (2024) smartphones.
Computations on such devices may be more in line with trajectory prediction application scenarios, providing a more valuable reference for efficiency analyses, rather than on high-performance server clusters.
Whether the batch size is 1 or 1000, it still could forecast trajectories within the \emph{low-latency} \cite{li2021spatial} thresholds, \IE, it could forecast trajectories within each adjacent sampling interval $\Delta t = 0.4$ seconds, even on the Apple M1 that performs similar to current iPhones.
Notably, its variation a2 (predict linear base + self-bias, see the original Tab. 4) takes only about 30\% inference time of the full model, still with acceptable quantitative performance, especially with better FDE than \EVMODEL~and \SCMODEL.
In addition, another variation, a8 (predict linear base + re-bias), owns the minimum number of parameters and also achieves better ADE than these baselines, indicating comparable prediction performance.
Therefore, depending on the application's needs, whether it is a fast calculation or an accurate one, the proposed \MODELSHORT~method may cope with it.

\section{Additional Discussions on Trajectory Biases}

\subsection{Contributions of Vibrations (Biases)}

\begin{figure}[t]
    \centering
    \includegraphics[width=1.0\linewidth]{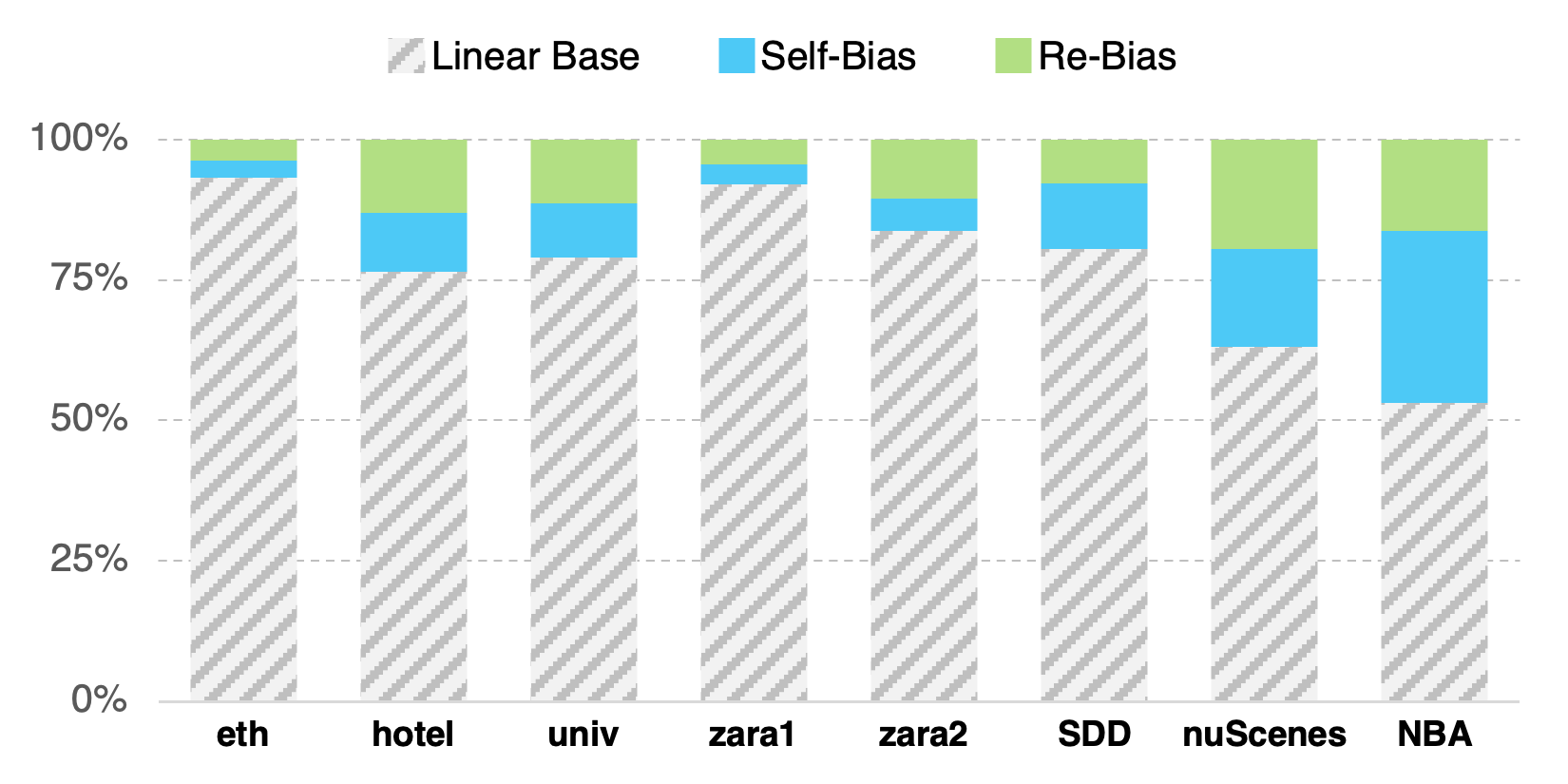}
    \caption{
        Energy shares of trajectory biases on different datasets.
    }
    \label{fig_biasenergy}
\end{figure}

In the main manuscript, we have analyzed different trajectory biases and their spatial distributions in a visualized way.
Here, we further discuss how these bias terms contribute quantitatively to final forecasted trajectories.
We use the energy of trajectories as a measurement to simplify the calculation.
It is computed as the square sum of each point in the trajectory $\mathbf{X}$, \IE, the $\|\mathbf{X}\|^2$.
\FIG{fig_biasenergy} reports these biases' average percentage energy shares across different datasets.
It can be seen that the linear base occupies the most energy, while other terms may change over datasets.
For example, the linear base occupies less energy on SDD, nuScenes, and NBA than those on ETH-UCY, indicating that there might be more non-linear trajectories, especially those caused by socially interactive behaviors.

This phenomenon aligns with previous research \cite{makansi2021you} that the simulation of trajectories themselves (the non-interactive terms, \IE, the linear base plus the self-bias in the proposed \MODELSHORT) has become the most important optimization objective for most trajectory prediction networks.
Under this circumstance, we notice that the re-bias (under social excitations) has not been compressed significantly, particularly in comparison to the self-bias (Newtonian excitations).
Instead, self-bias and re-bias provide almost identical contributions.
This implies that the predicted trajectories have been decomposed into distinct biases that hold approximately the same energy (instead of being drowned out by other terms) except for their linear portions, thus roughly validating the usefulness of these biases.

\subsection{Spatial Distributions of Vibrations}

\begin{figure}[t]
    \centering
    \includegraphics[width=1.0\linewidth]{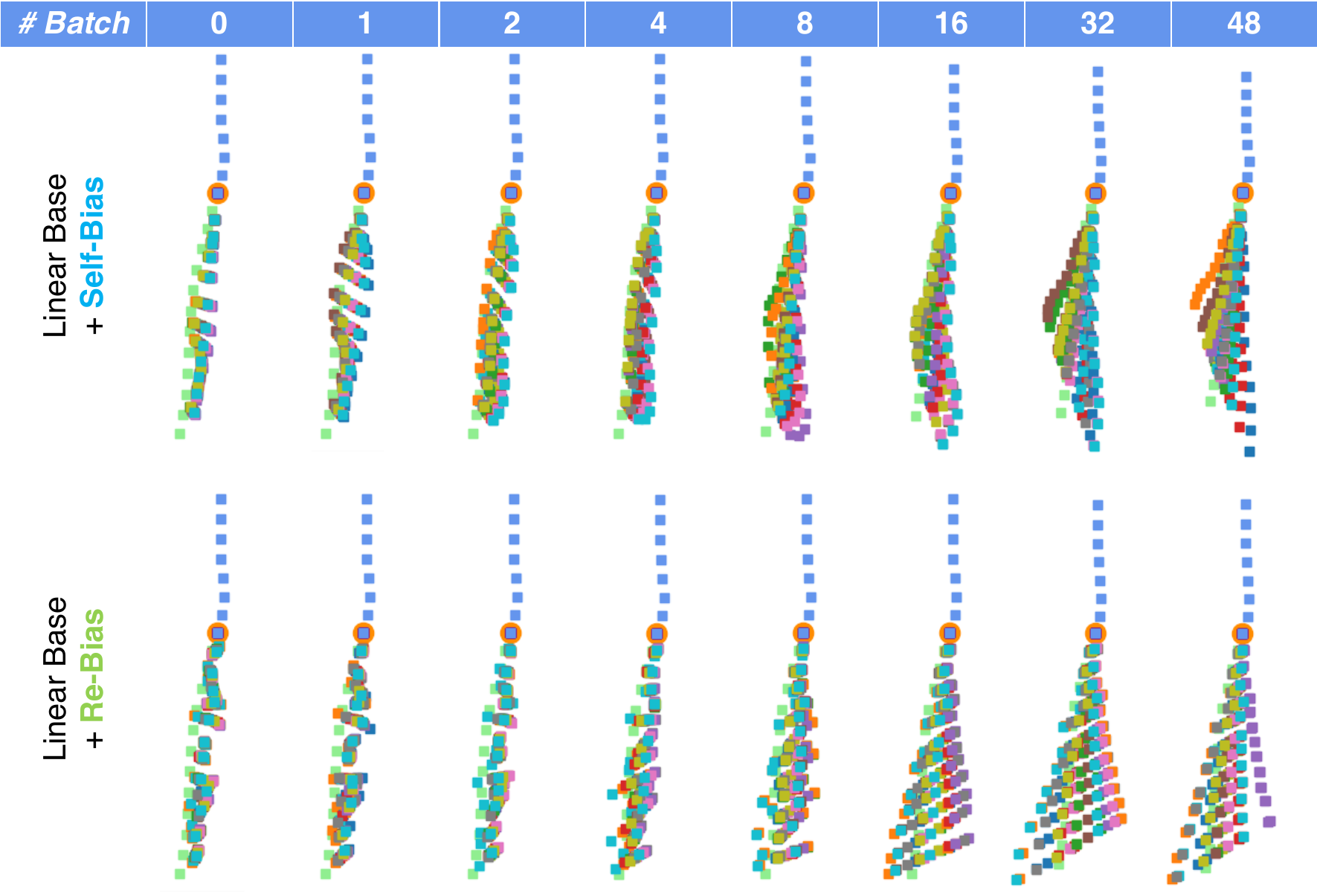}
    \caption{
        The changes of self-biases and re-biases during training (after how much training batches before finishing training 1 epoch of the whole dataset).
    }
    \label{fig_biaseswhentraining_minibatch}
\end{figure}

\begin{figure}[t]
    \centering
    \includegraphics[width=1.0\linewidth]{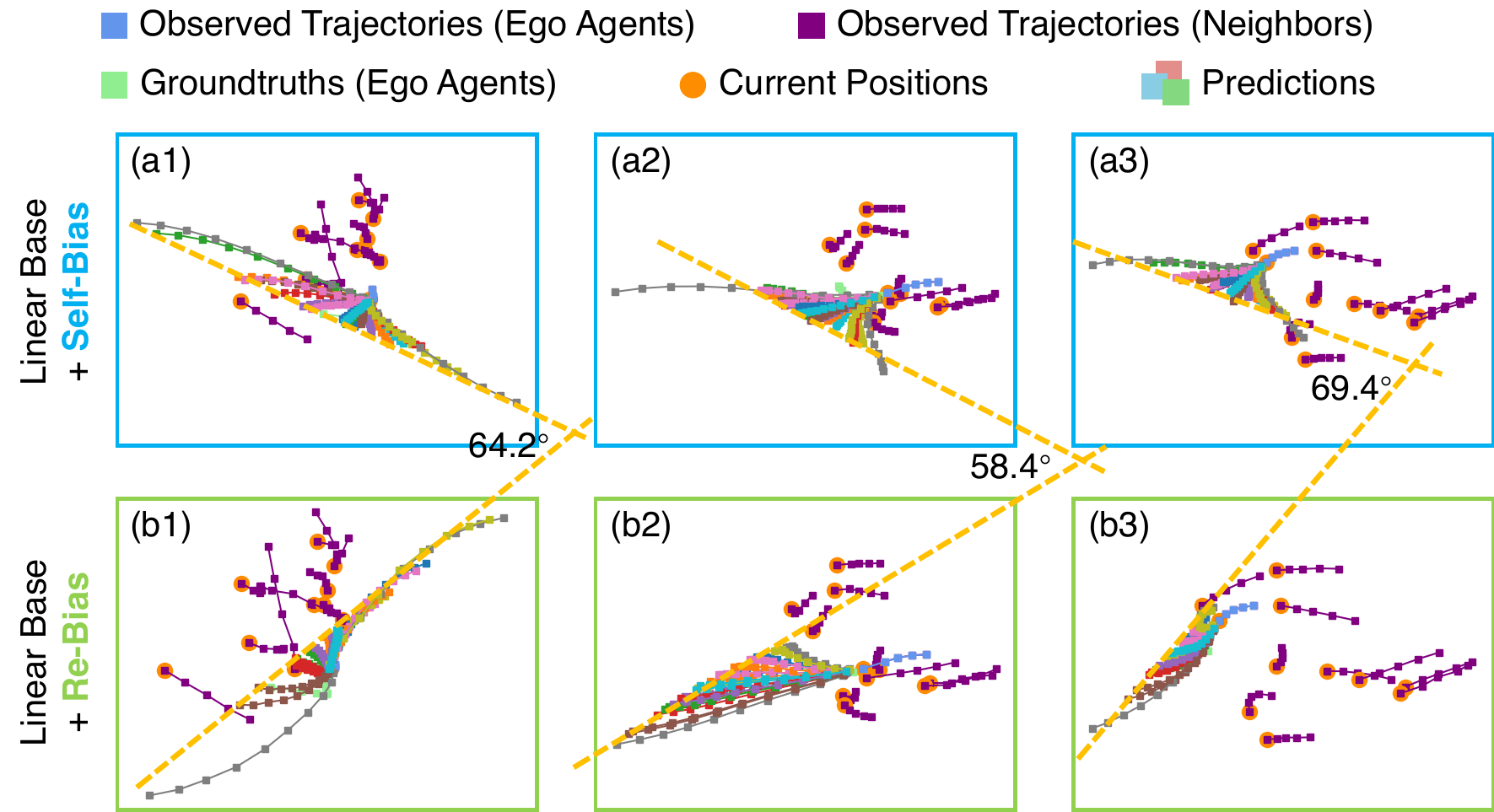}
    \caption{
        Visualized trajectory biases ($K=20$) on NBA.
        Different from other datasets, we found that self-biases and re-biases vibrate in directions about $\pi/3$ (60 degrees) away from each other.
    }
    \label{fig_biases_nba}
\end{figure}

\begin{figure*}[t]
    \centering
    \includegraphics[width=1.0\linewidth]{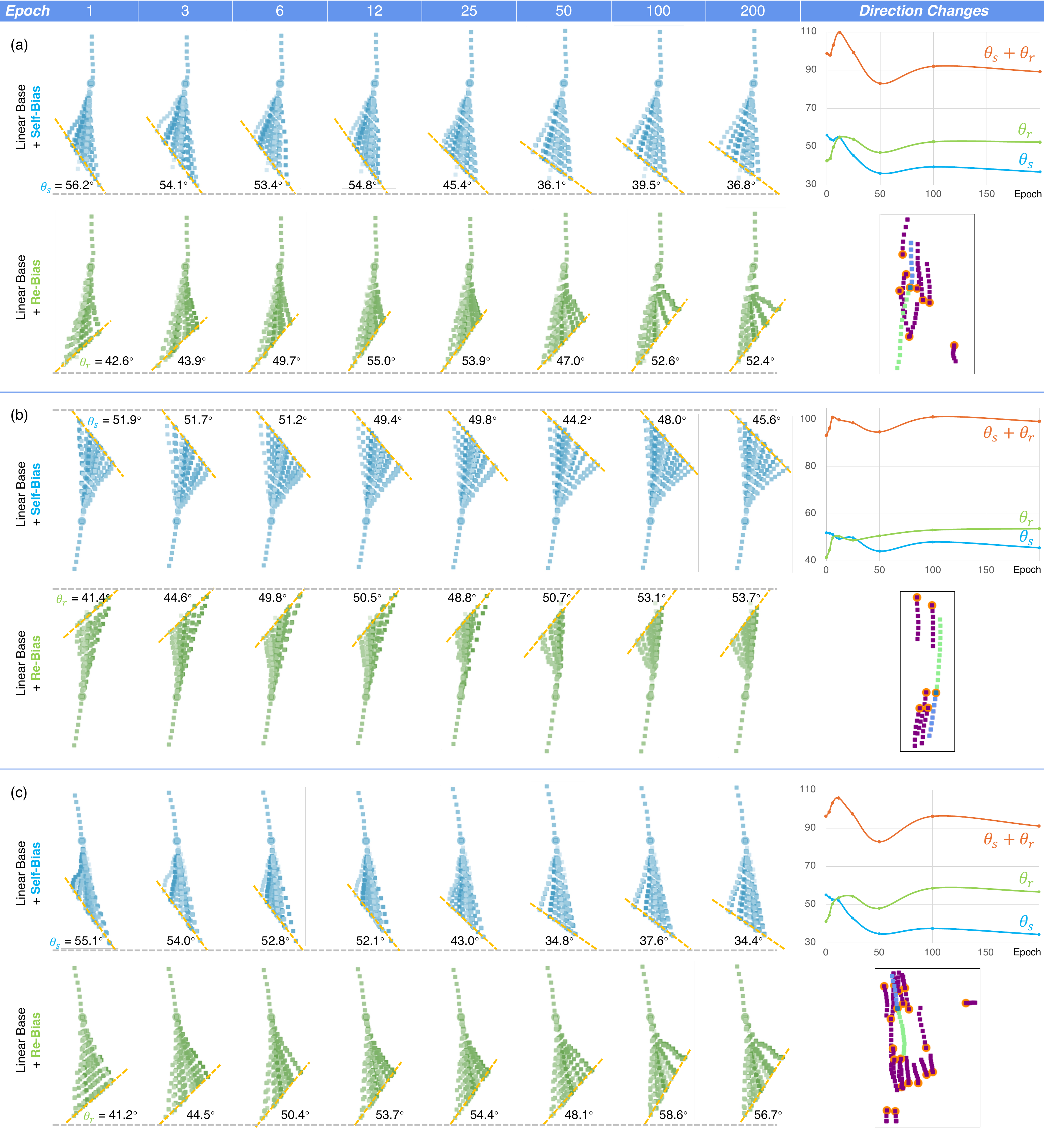}
    \caption{
        Vibration directions (defined as the acute angle between the fitting line of the predicted $K=20$ trajectory bias points on the last predicted moment and the horizontal direction, including the angle of self-biases $\theta_s$ and the angle of re-biases $\theta_r$) on different epochs when training on the pedestrian dataset ETH-UCY.
        We can consider self-bias and re-bias to be vibrating vertically when $\theta_s + \theta_r$ approaches $\pi/2$.
    }
    \label{fig_biaseswhentraining}
\end{figure*}

We have concluded in the main manuscript that self-bias and re-bias vibrate almost vertically in different directions.
We now further discuss this phenomenon.
Please note that this phenomenon is different from our angle-based approach to gather resonance features when forecasting re-biases, even though words like ``angles'' or ``directions'' are evolved in both these approaches.
Such a vertically-vibrating phenomenon is directly caused by our \emph{vibration-like} trajectory prediction strategy that forecasts trajectories with multiple trajectory biases.
For the convenience of representation, we define the \emph{vibration direction} of a trajectory bias as the acute angle between the fitting line of the predicted $K=20$ trajectory bias points on the last predicted moment and the horizontal direction.
Thus, we have angles $\theta_s$ and $\theta_r$ to represent how self-biases and re-biases vibrate.

\FIG{fig_biaseswhentraining} shows how angles $\theta_s$ and $\theta_r$ change with training epochs during one training progress.
It can be seen that each angle, whether $\theta_s$ or $\theta_r$, gradually converges to a particular value for each sample during training.
These convergence processes may be followed by numerical oscillations, and the convergence values may not be the same for every sample.
Especially, we observe that the convergence value of $\theta_s + \theta_r$ is around $\pi/2$ (90 degrees) for most pedestrian samples, which is our described vertically-vibrating phenomenon in the main manuscript.
In addition, it also shows that the nonlinearities of these trajectory biases have been reassigned during training.
For example, we can see stronger nonlinearities in the predicted self-biases after 3 training epochs than those in re-biases in \FIG{fig_biaseswhentraining} (a) and (c).
However, self-biases become more linearly distributed over time (as expected when wiring the Transformer $T_s$).
Such nonlinearities seem to be \emph{taken over} by the other re-biases, better seen by comparing the forecasted re-biases at epochs 50 to 200 in \FIG{fig_biaseswhentraining} (a) and (c).

\FIG{fig_biaseswhentraining_minibatch} represents how these trajectory biases are distributed and how they change during training at several initial training steps (after how many batches of training data).
It can be seen that these biases are almost randomly distributed at the very first training step, with none of the linearity/nonlinearity or vertically vibrating properties presented.
We also observe that it is the re-bias that leads the training.
For example, the phenomenon that predicted points in re-biases distributed around a straight line has already appeared after 16 batches in \FIG{fig_biaseswhentraining_minibatch}, while it appears for self-biases until training after 32 batches.

In addition to our descriptions of the original Fig. 8 in the main manuscript, the changes in these biases can also be roughly explained as a two-player cooperative game.
In detail, our goal is to learn to forecast two trajectory biases (the linear base is not trainable), $\Delta \hat{\mathbf{Y}}^i_s$ and $\Delta \hat{\mathbf{Y}}^i_r$, to fit the given trajectory ${\mathbf{Y}}^i - \hat{\mathbf{Y}}^i_l$.
Also, these two biases share almost equal positions in the prediction network, since whether their mirrored Transformer structures or they are exactly the same additive weight (both equal to 1, since $\hat{\mathbf{Y}}^i = \hat{\mathbf{Y}}^i_l + \Delta \hat{\mathbf{Y}}^i_s + \Delta \hat{\mathbf{Y}}^i_r$).
Intuitively, these biases would not have the same amplitude and would not vibrate in the same direction, as this would cause the whole network to get confused and not be able to distinguish samples that have similar ego trajectories but with different neighbors and different future trajectories, since self-biases are learned and predicted only by the ego agent, whereas re-biases are by the ego agent and all neighbors together.
In other words, only re-biases could \emph{tell} such socially different differences.
This means that these two biases are more inclined to vibrate in different directions, and re-biases would be the first ones to distinguish different training samples, where the differences of their future trajectories are mostly caused by their differently distributed neighbors.
Under this circumstance, the differences between self-biases and re-biases may increase as the training progresses.
Thus, after a period of cooperation, these biases would likely show the greatest difference in vibrational directions, leading to our observed vertically vibrating phenomenon.

Note that these explanations are only meant to provide a more intuitive understanding of two trajectory biases vibrating vertically to each other and are not strict proofs.
After our validation, similar phenomena have been presented on pedestrian datasets ETH-UCY and SDD.
However, as we show in the original Fig. 7 (a6) and (c6) in the main manuscript, most NBA samples do not match this rule, with about 60 degrees of direction differences between these two trajectory biases.
We attempted to explain this phenomenon by assuming that in NBA scenarios, where linear bases are difficult to apply (due to the rapidly changing motion states of the players and the intent of the game), there may actually be three \emph{actual} trajectory biases, and thus the best training is achieved when these three biases vibrate at a 60-degree difference from each other.
However, two \emph{actual} biases were estimated simultaneously in self-biases or re-biases, thus causing this vibration phenomenon with a 60-degree difference between each other (as shown in \FIG{fig_biases_nba}).
This phenomenon still needs further study.

\section{Further Discussions on Resonance Features and Social Interactions}

\subsection{Distributions of Resonance Features}

\begin{figure}[t]
    \centering
    \includegraphics[width=1.0\linewidth]{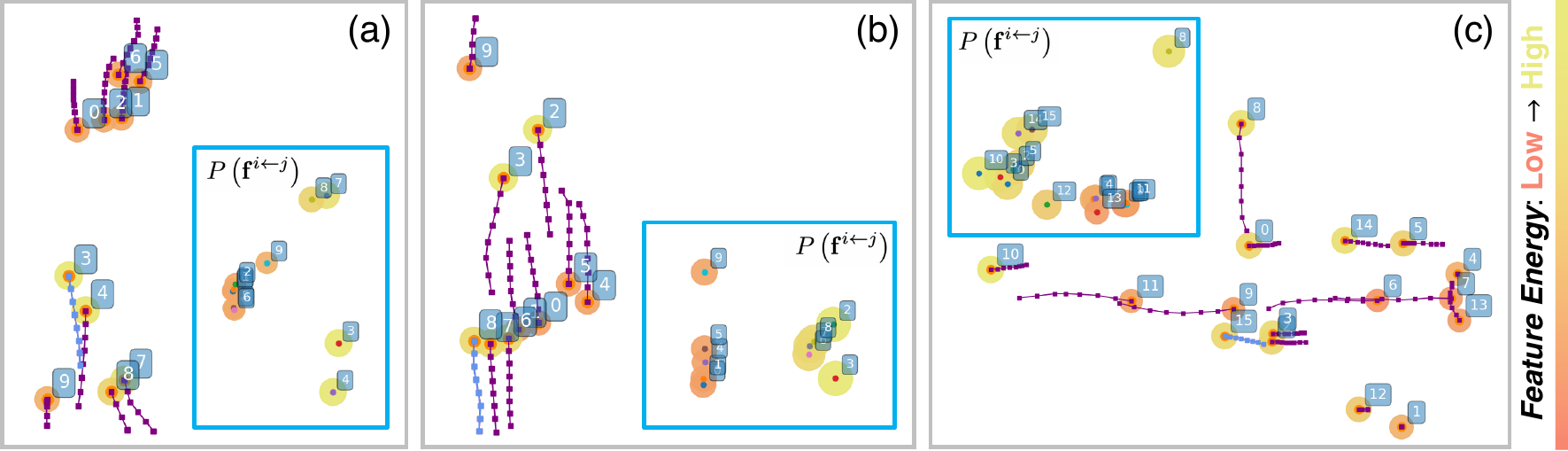}
    \caption{
        Further comparisons of resonance features in more complex interactive scenes.
        Colored dots represent the value of feature energy $\|\mathbf{f}^{i \leftarrow j}\|^2$, and each blue box represents resonance features in the feature space $\{\mathbf{f}^{i \leftarrow j} | 1\leq j \leq N_a, j \neq i\}$.
    }
    \label{fig_resonance2}
\end{figure}

In the main manuscript, we have analyzed resonance features of different neighboring agents $j$ relative to the ego $i$ and have observed a clear tendency to describe all neighbors in distinct groups.
However, analyzing high-dimensional features using the energy metric results in a large amount of information loss and only partially reflects their nature.
In \FIG{fig_resonance2}, we use PCA (Principal Component Analysis) to reduce the dimension of each $\mathbf{f}^{i \leftarrow j}$, thus further visualizing how they distribute in the feature space.
It can be seen that agents with similar motion states share closer features.
Meanwhile, agents with similar features may probably behave as groups in real scenarios, like the group \{0, 1, 2, 5, 6\} in \FIG{fig_resonance2} (a) and \{0, 1, 4, 5\} in \FIG{fig_resonance2} (b), independent of their relevant positions (since the relative positional information has been detached in these features, see the original Eq. (9)).
It is worth noting whether or not following a group is actually an important social event and has been studied a lot by previous researchers like GroupNet \cite{xu2022groupnet}.
The proposed \MODELSHORT~could learn such behaviors without constructing graph structures and human annotations, further verifying our thought that social interactions are associated with the spectral properties of trajectories.

Also, we can see from \FIG{fig_resonance2} that agents with different preferences could be distinguished, like bikers \{6, 7, 9, 11\} and \{8\} owns remarkable differences to other pedestrians in \FIG{fig_resonance2} (c).
This means that the proposed \MODELSHORT~has the ability to learn different social responses of the same ego agent to neighbors with different states.
It also indicates that \MODELSHORT~has learned social behaviors that are not limited to finding or locating which neighbors have similar spectral characteristics to itself, but focuses more on those with different spectrums to itself, like the feature distances between group \{0, 1, 2, 5, 6\} and others in the feature space in \FIG{fig_resonance2} (a).
Thus, our assumptions about the \emph{social} resonance can be verified, which regards that social interactions are associated with the spectral properties of trajectories, and it could produce the maximum effects when the trajectory spectrums of a neighbor and the ego agent both cover some common frequency bands (\IE, higher spectral similarities) or no spectral overlap at all (\IE, have a longer feature distance in the feature space).

\subsection{Resonance Features and Neighbor Positions}

\begin{figure*}[t]
    \centering
    \includegraphics[width=1.0\linewidth]{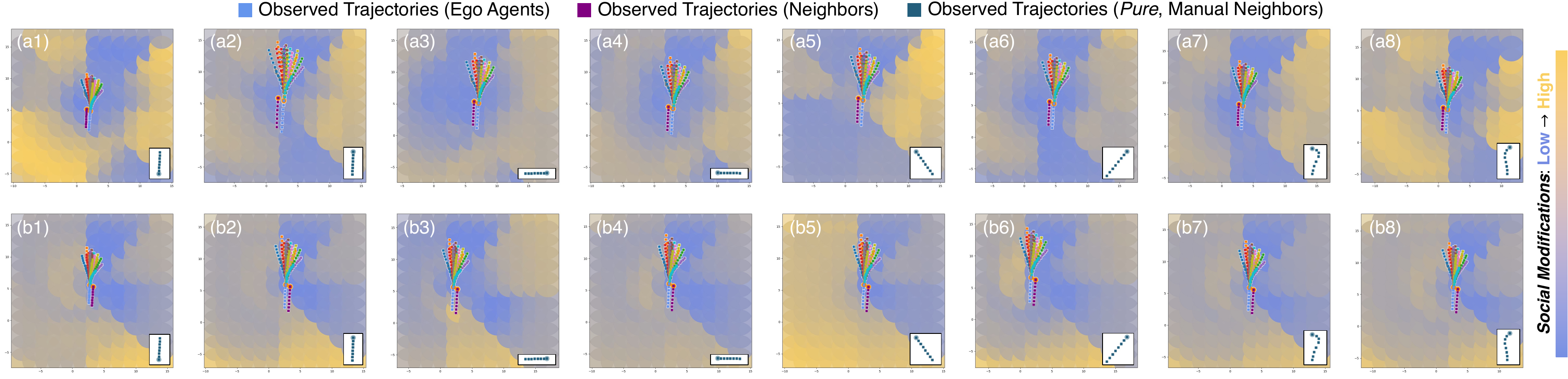}
    \caption{
        Spatial distribution $P(c(x, y | \mathbf{X}^m))$ of more structured of manual neighbors of the proposed \MODELSHORT~model.
    }
    \label{fig_moretraj}
\end{figure*}

\begin{figure}[t]
    \centering
    \includegraphics[width=1.0\linewidth]{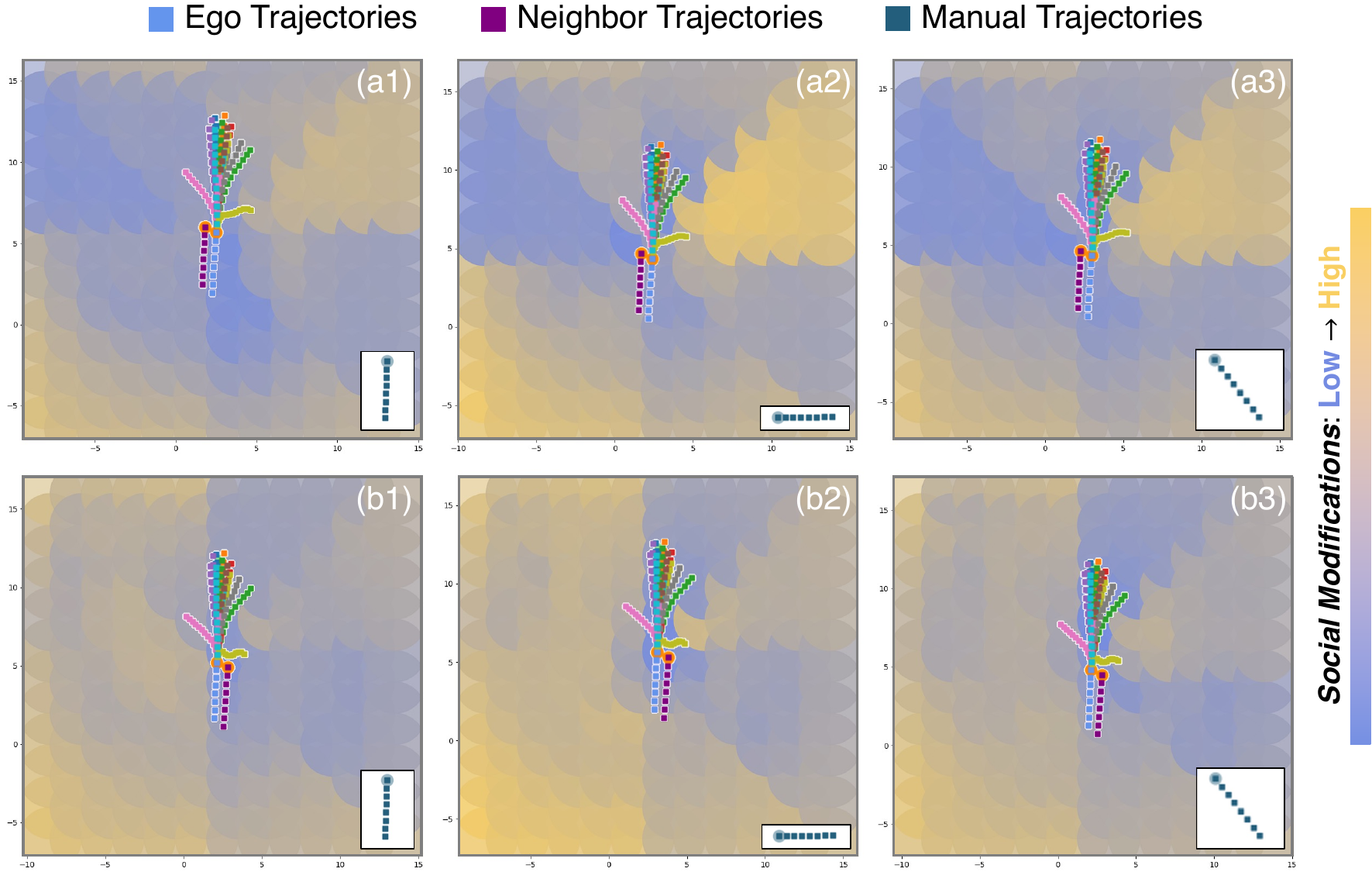}
    \caption{
        Spatial distribution $P(c(x, y | \mathbf{X}^m))$ of more structured of manual neighbors of \SCMODEL\SCCITE~model.
    }
    \label{fig_moretraj_sc}
\end{figure}

As described in the original Eq. (9), the relative positional information of all neighbors relative to the ego agent has not been considered in resonance features.
In other words, we regard the resonance features to represent the \emph{pure} spectral properties of agents' trajectories, since it can easily be verified that a positional offset introduces more low-frequency (or base frequency) interference into the corresponding spectrum for the linear transform $\mathcal{T}$ we used.
Instead, in the original Eq. (11), we encode the corresponding positional information outside from the resonance feature $\mathbf{f}^{i \leftarrow j}$ as the $\mathbf{f}^{i \leftarrow j}_p$.
In our main manuscript, we have used term $c(x, y | \mathbf{X}^m)$ to verify how the resonance feature and the positional information collaborate to \emph{modify} the final predicted trajectories:

Define the set of observed trajectories of all agents as $\mathcal{X} = \{\mathbf{X}^i | 1 \leq i \leq N_a\}$, and the computation of a prediction network as $\mathcal{N}(\cdot)$, its prediction for the $i$th ego agent can be represented by
\begin{equation}
    \hat{\mathbf{Y}}^i := \left(\hat{\mathbf{p}}^i_{t_h + 1}, ..., \hat{\mathbf{p}}^i_{t_h + t_f}\right)^\top = \mathcal{N} (\mathcal{X}).
\end{equation}
Considering a manual neighbor \SCCITE, \IE, a neighbor with manually set observed trajectories and has been put manually into the prediction scene, located at $(x, y)$ with the \emph{pure} trajectory $\mathbf{X}^m$, we denote the newly predicted trajectory after applying the \emph{intervention} $do(\overline{\mathcal{X}} = \mathcal{X} \cup \{\mathbf{X}^m + (x, y)^\top\})$ as
\begin{equation}
    \overline{\hat{\mathbf{Y}}^i} := \left(\overline{\hat{\mathbf{p}}}^i_{t_h + 1}, ..., \overline{\hat{\mathbf{p}}}^i_{t_h + t_f}\right)^\top = \mathcal{N} (\overline{\mathcal{X}}).
\end{equation}
Then, the absolute trajectory modification $c(x, y| \mathbf{X}^m)$ caused by the new trajectory $\mathbf{X}^m + (x, y)^\top$ is defined as
\begin{equation}
    c(x, y| \mathbf{X}^m) = \max_{t_h + 1 \leq t \leq t_h + t_f} \left\|\overline{\hat{\mathbf{p}}}^i_t - \hat{\mathbf{p}}^i_t \right\|.
\end{equation}
This is actually an intervention to verify the model's ability to represent \emph{causalities} \cite{granger1969investigating} between social interactions and predicted trajectories.
If the model cannot represent such a causal relationship, then the predicted trajectories may not change no matter how much the new neighbor changes.

Such modification describes how the \emph{originally} predicted trajectories are modified by additionally considering a new neighbor agent and potential socially interactive behaviors with it.
For clarity, we abbreviate such modifications as \emph{social modifications} in the following sections.
Thus, by manually moving (translating) the trajectory $\mathbf{X}^m$ to any position $(x, y)$ around the ego agent, we can verify how the resonance feature describes a neighbor with different positions relative to the ego agent.
Spatial distributions of $c(x, y| \mathbf{X}^m)$ corresponding to more structured manual trajectories ($\mathbf{X}^m$) are attached in \FIG{fig_moretraj}, which represent how resonance features (how the manual trajectory is structured) and positional information (where the manual neighbor is located, and its relative position to other neighbors) collaborate to modify the original forecasted trajectories.

\subsection{Additional Discussions on Social Modifications}

The above trajectory modification $c(x, y| \mathbf{X}^m)$ could represent how a trajectory prediction model responds to change its predictions as a consideration of social interactions.
Naturally, the representation of social behaviors can be considered better when such social modifications of a model are more sensitive and consistent with human intuitions.
\FIG{fig_moretraj_sc} presents the spatial distribution of social modifications $c(x, y| \mathbf{X}^m)$ obtained from one of the current state-of-the-art approaches \SCMODEL\SCCITE, which is proposed to mainly focus on social interactions among agents.

Comparing \FIG{fig_moretraj_sc} and \FIG{fig_moretraj}, it can be seen that \SCMODEL~fails at capturing the state of this manual neighbor, especially how its trajectory is structured.
For example, the spatial distribution of social modifications is almost the same across \FIG{fig_moretraj_sc} (a1), (a2), and (a3), even though the trajectories of these manual neighbors are totally different.
On the contrary, the proposed \MODELSHORT~could better capture their differences through the resonance features, better seen by comparing \FIG{fig_moretraj} (b2) (b5) \emph{v.s.} \FIG{fig_moretraj_sc} (b1) (b3) that share the same prediction situation.
It can be seen that the former pair present totally different social responses, especially to the left side of the ego agent, while the other pair shares almost the same spatial distribution.
In addition, the differences of social modifications on different ego agents provided by the \SCMODEL~is also relatively small, while those \MODELSHORT~ones could better describe such interaction differences, better seen by comparing \FIG{fig_moretraj} (a5) and (b5) that present an almost completely opposite spatial distribution.
Therefore, we can conclude that the proposed \MODELSHORT~could better capture the fine-level interaction differences among different egos and neighbors, thus forecasting better social-aware trajectories.

\begin{table*}[tbph]
    \centering
    \footnotesize
    \begin{tabular}{lcccccc}
        \toprule
        ID & $\mathcal{T}$ & eth & hotel & univ & zara1 & zara2 \\

        \midrule
        b1 & DFT & 0.240/{0.359} & 0.112/{0.174} & 0.258/0.459 & 0.181/0.306 & {0.137/0.237} \\
        b2 & Haar (default) & \C{0.232/0.354} & \C{0.103/0.152} & \C{0.242/0.414} & \C{0.172/0.290} & \C{0.131/0.225} \\
        b3 & None & {0.238}/0.364 & {0.111}/0.178 & {0.248/0.436} & {0.177/0.302} & 0.138/0.237 \\

        \bottomrule
    \end{tabular}
    \caption{
        Ablation studies on transforms used to obtain trajectory spectrums, including Haar (default), DFT, and none transform.
    }
    \label{tab_transform}
  \end{table*}

\begin{table*}[tbph]
    \centering
    \footnotesize
    \begin{tabular}{lccccccc}
        \toprule
        ID & $N_{way}$ & eth & hotel & univ & zara1 & zara2 \\

        \midrule
        c1 & 2 & 0.249/0.384 & 0.104/0.153 & 0.276/0.475 & 0.181/0.301 & 0.137/0.234 \\
        c2 & 3 & 0.236/0.362 & \C{0.102/0.150} & 0.253/0.446 & \C{0.178/0.299} & \C{0.131/0.223} \\
        c3 & 4 (default) & \C{0.232/0.354} & \C{0.103/0.152} & \C{0.242/0.414} & \C{0.172/0.290} & \C{0.131/0.225} \\
        c4 & 6 & \C{0.229/0.356} & \C{0.102}/0.154 & 0.250/0.441 & 0.182/0.310 & 0.135/0.229 \\
        c5 & 12 & 0.237/0.378 & 0.104/0.155 & \C{0.245/0.437} & 0.180/0.307 & 0.134/0.232 \\

        \bottomrule
    \end{tabular}
    \caption{
        Ablation studies on the number of waypoints $N_{way}$ when forecasting self-biases.
    }
    \label{tab_waypoints}
  \end{table*}

\begin{table*}[tbp]
    \centering
    \footnotesize
    \begin{tabular}{lcccccc}
        \toprule
        ID & Interaction Representation & eth & hotel & univ & zara1 & zara2 \\

        \midrule
        d1 & Social Pooling & 0.243/0.376 & 0.107/0.166 & 0.256/0.461 & 0.183/0.312 & 0.138/0.242 \\
        d2 & GCN & 0.240/0.272 & 0.107/0.164 & 0.253/0.448 & 0.179/0.311 & 0.133/0.230 \\
        d3 & \SCMODEL & 0.238/0.370 & 0.112/0.181 & 0.252/0.444 & 0.180/0.303 & 0.139/0.241 \\
        d4 & Resonance Gathering (Ours) & \C{0.232/0.354} & \C{0.102/0.152} & \C{0.242/0.414} & \C{0.172/0.290} & \C{0.131/0.225} \\

        \bottomrule
    \end{tabular}
    \caption{
        Ablation studies on the usages of different social interaction representations when forecasting re-biases.
    }
    \label{tab_interaction}
  \end{table*}

\begin{table*}[tbp]
    \centering
    \footnotesize
    \begin{tabular}{lcccccc}
        \toprule
        ID & $N_\theta$ & eth & hotel & univ & zara1 & zara2 \\

        \midrule
        e1 & 1 & 0.237/0.363 & 0.102/\C{0.153} & 0.246/0.419 & 0.181/0.313 & \C{0.137}/0.238 \\
        e2 & 2 & \C{0.222}/0.391 & 0.103/0.156 & \C{0.246}/0.419 & 0.181/0.310 & 0.144/0.248 \\
        e3 & 4 & \C{0.232/0.350} & \C{0.101}/0.154 & 0.250/\C{0.416} & \C{0.179/0.303} & 0.139/\C{0.235} \\
        e4 & 8 (default) & \C{0.232/0.354} & \C{0.102/0.152} & \C{0.242/0.414} & \C{0.172/0.290} & \C{0.131/0.225} \\
        e5 & 12 & 0.249/0.390 & 0.103/0.159 & 0.254/0.426 & 0.178/0.299 & 0.145/0.250 \\
        e6 & 16 & 0.253/0.391 & 0.104/0.160 & 0.252/0.429 & 0.178/0.295 & 0.146/0.247 \\

        \bottomrule
    \end{tabular}
    \caption{
        Ablation studies on the number of angle-based partitions $N_\theta$ when forecasting re-biases.
    }
    \label{tab_partitions}
  \end{table*}

\section{Additional Ablation Studies}

In the main manuscript, we have made ablation variations to verify the effect of different trajectory biases in the proposed \MODELSHORT.
However, due to the page limitations, some other hyperparameters, like the number of waypoints $N_{way}$, the number of angle partitions $N_\theta$, or the transform $\mathcal{T}$, have not been validated, since they are set following the experimental results of former researches \cite{wong2023another,wong2023socialcircle,wong2022view}.
Here, we conduct several ablation variations and report their corresponding ablation results.
Please note that these validations are only to determine the parameter choices since these parts are not the main contributions of the proposed \MODELSHORT.

\subsection{Ablation: Spectrums and Transform Types}

Our core idea is to forecast trajectories as different vibration portions and regard that social interactions are associated with the spectral properties of agents' trajectories.
Naturally, we need to use some transform to get trajectory spectrums.
The Fourier transform and its variations have been used in a wide variety of fields.
\EVMODEL~\EVCITE~applies discrete Fourier transform (DFT) and discrete Haar transform to obtain trajectory spectrums so that the trajectories can be predicted hierarchically via different spectral components.

According to their experimental results, Haar transform has better time efficiency than DFT, and it could also better describe signals with rapidly changing characteristics (with a relatively lower \emph{Vanishing Moment}).
Thus, we choose Haar transform to obtain trajectory spectrums in this manuscript.
We have also conducted several ablation experiments to verify its usefulness compared to the non-transform ones and the DFT, whose results are reported in \TABLE{tab_transform}.
We can see that the Haar variation b2 obtains the best performance across these ETH-UCY sets, compared to whether DFT variation b1 or the non-transform variation b3.
Especially, such performance enhancements appear to be more effective on FDE, with about 6.5\% better average FDE than variation b1, and about 5.4\% better than variation b3.
Such experimental results verify the usefulness of the Haar transform.
Since we are not the first to work on forecasting trajectories with Haar transforms, we think this part of the analysis is not the necessary for the main manuscript.

\subsection{Ablation: the Number of Waypoints $N_{way}$}

We regard self-bias as roughly reflecting agents' intention changes or random behaviors on limited waypoints rather than the whole prediction period.
Our idea is to reduce the computational load by such an operation and improve the network's generalization to prevent it from overfitting to some training samples when forecasting agents' random behaviors.
According to previous research \cite{mangalam2020s,wong2022view}, a too-low waypoint setting may lead to the loss of the accuracy of intention prediction, and conversely, a too-high setting may limit the ability of prediction models to forecast stochastic trajectories.
\TABLE{tab_waypoints} reports the ablation results on changing the $N_{way}$.
It can be seen that \MODELSHORT~obtains the best average performance when setting $N_{way}$ to 4.
Also, the usage of waypoints is not our main contribution.
Thus, this ablation table is not included in the main manuscript.

\subsection{Ablation: Social-Interaction-Representation}

One significant concern is representing agents' interaction context when making predictions.
Its core is to properly share motion status with all neighboring agents.
In \MODELSHORT, we use the resonance feature $\mathbf{f}^{i \leftarrow j}$ to represent the single-directional relationship caused by a neighbor $j$ onto the ego agent $i$.
Then, the angle-based resonance gathering is used to gather features into the ego agent as the final social interaction representation, \IE, the resonance matrix $\mathbf{F}^i_{\mathcal{R}}$.
This information-gathering approach is inspired by the previous work \SCMODEL~\SCCITE.
We have also conducted ablation studies to verify its effectiveness to other social-interaction-representation approaches, including the Social Pooling \cite{alahi2016social}, graph convolutional network \cite{kipf2016semi}, and the vanilla \SCMODEL.
Results are reported in \TABLE{tab_interaction}.

\subsection{Ablation: the Number of Angle Partitions $N_\theta$}

Like \SCMODEL~\SCCITE, we use an angle-based resonance gathering method to gather all neighbors' resonance features in each angle-based partition.
According to their experimental results, setting $N_\theta = t_h$ may obtain the best prediction performance.
Here, we conduct ablation variations to verify how the number of partitions $N_\theta$ affects model performance.
Results are reported in \TABLE{tab_partitions}.
We observe that \MODELSHORT~obtains the best performance when setting $N_\theta = 8$ ($=t_h$), indicating similar conclusions of angle-based partitions in \SCMODEL.
Also, the angle-based partitioning is only used to gather resonance features, which is not the main concern of this manuscript.
Thus, these results are only used as a reference for parameter selection.

\section{Further Discussions on the Vehicle Dataset}
\label{sec_vehicle}

\begin{table}[tbp]
    \centering
    \footnotesize
    \begin{tabular}{lcc}
        \toprule
        Method (nuScenes) & \emph{best-of-5} $\downarrow$ & \emph{best-of-10} $\downarrow$ \\

        \midrule
        Trajectron++\TRAJECTRONPPCITE~(2020) & 3.14/7.45 & 2.46/5.65 \\
        \YNET\YNETCITE~(2020) & 2.46/5.15 & 1.88/3.47 \\
        \SOPERMODEL\SOPERMODELCITE~(2025) & 2.21/4.58 & 1.79/3.40 \\
        AgentFormer\AGENTFORMERCITE~(2021) & 1.86/3.89 & 1.45/2.86 \\
        \DICE\DICECITE~(2024) & 1.76/3.70 & 1.44/2.67 \\
        \AFFLN\AFFLNCITE~(2024) & 1.83/3.78 & 1.32/2.73 \\
        \EVMODEL\EVCITE~(2023) & 1.46/3.18 & 1.15/2.37 \\
        \SCMODEL\SCCITE~(2024) & \C{1.44/3.10} & \C{1.13/2.30} \\
        \MUSEVAE\MUSEVAECITE~(2022) & \C{1.38/2.90} & \C{1.09/2.10} \\

        \midrule
        \MODELSHORT~(Ours) & \C{1.26/2.72} & \C{1.03/2.12} \\

        \bottomrule
    \end{tabular}
    \caption{
        Comparisons to other state-of-the-art methods on nuScenes.
        Metrics reported are ``ADE/FDE'' in meters under \emph{best-of-5} and \emph{best-of-10}.
    }
    \label{tab_app_sota_nuscenes}
    \end{table}

\begin{table*}[tbph]
    \centering
    \footnotesize
    \begin{tabular}{ccc}
        \toprule
        ID & ~$l$~~$\Delta_s$~$\Delta_r$ &
        nuScenes $\downarrow$ \\

        \midrule
        a1 & \cm~~\xm~\xm       &  3.48/7.93         \\
        a2 & \cm~~\cm~\xm       &  1.26/2.36         \\
        a3 & \xm~~\xm~~SC       &  1.09/2.30         \\
        a4 & \xm~~\xm~~Re       &  \C{1.04/2.16}     \\
        a5 & \xm~~\cm~~SC       &  1.06/2.24         \\
        \bottomrule
    \end{tabular}
    \begin{tabular}{ccc}
        \toprule
        ID & ~$l$~~$\Delta_s$~$\Delta_r$ &
        nuScenes $\downarrow$ \\

        \midrule
        a6 & \xm~~\cm~~Re       &  \C{1.03/2.12}     \\
        a7 & \cm~~\xm~~SC       &  1.13/2.45         \\
        a8 & \cm~~\xm~~Re       &  1.13/2.44         \\
        a9 & \cm~~\cm~~SC       &  1.06/2.23         \\

        a0 (default) & \cm~~\cm~~Re & \X{1.05/2.17} \\
        \bottomrule
    \end{tabular}

    \caption{
        Ablation studies on nuScenes under \emph{best-of-10}.
        ``$l$'', ``$\Delta_s$'', ``$\Delta_r$'' represent whether $\hat{\mathbf{Y}}^i_l$, $\Delta \hat{\mathbf{Y}}^i_s$, and $\Delta \hat{\mathbf{Y}}^i_r$ are superposed when training.
        ``Re'' and ``SC'' denote resonance gathering and \SCMODEL~\SCPCITE~when learning $\mathbf{V}^i_r$.
        Results marked in \X{Red} denote worse ones than the best variation.
    }
    \label{tab_app_ab_nuscenes}
  \end{table*}

\begin{figure*}[t]
    \centering
    \includegraphics[width=1.0\linewidth]{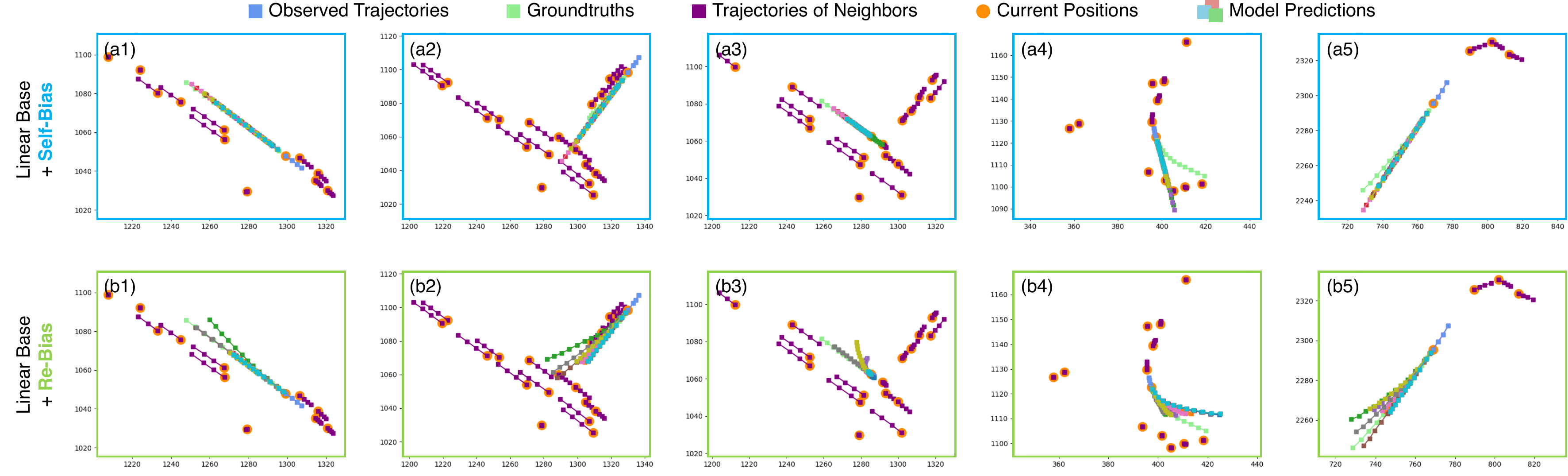}
    \caption{
        Visualized trajectory biases ($K=10$) on nuScenes.
    }
    \label{fig_biases_nuscenes}
\end{figure*}

The proposed \MODELSHORT~model is designed to forecast pedestrian trajectories.
Furthermore, we now discuss and analyze its prediction performance quantitatively and qualitatively on the large-scale vehicle dataset \textbf{nuScenes} \cite{caesar2020nuscenes,caesar2019nuscenes} collected in urban cities.
We split 550/150/150 scenes to train/test/val only on vehicles, under $\left(t_h, t_f, \Delta t\right) = (4, 12, 0.5)$ \cite{lee2022muse,saadatnejad2022pedestrian}.
This section further discusses how \MODELSHORT~works to forecast vehicle trajectories, especially the usefulness of the \emph{vibration-like} trajectory prediction strategy.

\subsection{Quantitative Analyses}
As shown in \TABLE{tab_app_sota_nuscenes}, we observe that \MODELSHORT~achieves an impressive performance compared to the state-of-the-art methods.
Especially, \MODELSHORT~outperforms the newly published model \SOPERMODEL~by 42.9\%/40.6\% ADE/FDE when generating 5 trajectories while the improvement in FDE even reaches as high as 60.3\% when generating 10 trajectories.
In addition, compared to the pedestrian-focused \SCMODEL, \MODELSHORT~obtains about 8.8\%/7.8\% ADE/FDE improvements when generating 10 trajectories.
Nevertheless, \MODELSHORT~loses about 1.0\% FDE compared to the state-of-the-art vehicle-centric \MUSEVAE, but it obtains better performance when only generating 5 fewer trajectories for each ego agent, with about 8.7\%/6.2\% better performance.
Also, in \TABLE{tab_app_ab_nuscenes}, variations present similar trends like those on NBA, reported the original Tab. 4, that \MODELSHORT~perform even worse when adding linear bases in the final predictions.
Variations a7 to a9 exhibit performance drops of up to 7.6\%/12.9\%, which are rarely seen in ETH-UCY and SDD.
Nevertheless, vibrations (self-biases and re-biases) could still help to improve the performance (variation a6), especially compared to the \SCMODEL~variation a5.
Similar to our conclusion in the main manuscript, though the assumption that linear base serve as the final reference points may not apply to the vehicle scenes, vibrations and their sampled trajectory-biases still work continuously.
These results validate the competitiveness of \MODELSHORT, even in vehicle scenarios.

\subsection{Qualitative Analyses}
According to results in the original Figs. 7 and 8 in the main manuscript, or the above \FIG{fig_biaseswhentraining}, we conclude that self-biases are better at capturing intention changes or random path choices, represented as an additional vibration vertical to the direction of motion, while re-biases capturing social modifications as a vibration alongside the motion direction.
As shown in \FIG{fig_biases_nuscenes}, we observe that self-biases and re-biases vibrate in a different way when forecasting vehicle trajectories from those presented in pedestrian datasets.
It can be seen that self-biases now describe how ego vehicles' velocities would change, while re-biases describe whether they would turn in the future.
For example, \FIG{fig_biases_nuscenes} (a1) to (a5) indicate that different forecasted self-biases own different velocities (among $K=10$ random predictions), while they are almost distributed in the same direction, different from those in pedestrian cases.
Also, the forecasted re-biases indicate whether the vehicle will change its direction in the future (like to make a turn at an intersection in \FIG{fig_biases_nuscenes} (b2) and (b4)).
This phenomenon demonstrates the adaptability and effectiveness of the proposed vibration-like prediction strategy in capturing and predicting vehicle trajectories.
In particular, the difference in trajectory biases between pedestrian data and vehicle data further illustrates the adaptability of the proposed \MODELSHORT~to learn to fit trajectories according to the properties of ego agents.

Please note that the proposed \MODELSHORT~currently uses only observed trajectories (ego and neighbors) to predict future trajectories and does not include other kinds or modals of observations, which leads to its shortcomings in forecasting vehicle trajectories, like not being able to observe lane positions (which are already widely used by existing vehicle trajectory prediction methods \cite{wang2022ltp}), etc.
Therefore, although it shows the potential for vehicle trajectory prediction, the current approach is more applicable to pedestrian agents.
We will try to adapt it in the future to better suit vehicle prediction scenarios.


\end{document}